\documentclass{article} % For LaTeX2e
\usepackage{iclr2019_conference,times}

\usepackage{hyperref}
\usepackage{url}

\usepackage{amsmath}
\usepackage{amsfonts}
\usepackage{amsopn}
\usepackage{algorithm}
\usepackage{algpseudocode}
\DeclareMathOperator*{\argmin}{arg\,min}
\usepackage{graphicx}
\usepackage{subcaption}

\title{Practical Newton-Type Distributed Learning using Gradient Based Approximations}

\author{Samira Sheikhi \\
Department of Computer Science\\
University of Chicago\\
Chicago, IL 60637, USA \\
\texttt{\{ssheikhi\}@uchicago.edu}
}

\iclrfinalcopy % Uncomment for camera-ready version, but NOT for submission.
\begin{document}

\maketitle

\begin{abstract}
We study distributed algorithms for expected loss minimization where the datasets are large and have to be stored on different machines. Often we deal with minimizing the average of a set of convex functions where each function is the empirical risk of the corresponding part of the data. 
In the distributed setting where the individual data instances can be accessed only on the local machines, there would be a series of rounds of local computations followed by some communication among the machines. Since the cost of the communication is usually higher than the local machine computations, it is important to reduce it as much as possible. However, we should not allow this to make the computation too expensive to become a burden in practice. Using second-order methods could make the algorithms converge faster and decrease the amount of communication needed.

There are some successful attempts in developing distributed second-order methods. Although these methods have shown fast convergence, their local computation is expensive and could enjoy more improvement for practical uses. In this study we modify an existing approach, DANE (Distributed Approximate NEwton), in order to improve the computational cost while maintaining the accuracy. We tackle this problem by using iterative methods for solving the local subproblems approximately instead of providing exact solutions for each round of communication. We study how using different iterative methods affect the behavior of the algorithm and try to provide an appropriate tradeoff between the amount of local computation and the required amount of communication. Moreover, we use a subsample from the data on each machine for calculating an expensive part of the algorithm which is calculating the full gradient. Our experiments show using subsamples of data for this part does not have a big influence on the overall behavior with our algorithms. We use this property to provide solution for solving learning problems in streaming applications. We demonstrate the practicality of our algorithm and compare it to the existing distributed gradient based methods such as SGD.
\end{abstract}

\section{Introduction}

\subsection{Motivation and importance of distributed learning}
With the exceeding amount of data collected by inexpensive information sensing-devices, cameras, microphones, or the internet, etc. and in parallel the availability of higher capacities for storing information, new challenges are introduced in the field of machine learning. There is a critical need to adapt and scale the existing machine learning algorithms to this new distributed setting. In particular, storing and analyzing the data on a single machine would not be possible in many cases because of limited storage capacity and processing power on a single machine, the expensive cost of transferring data, and unwillingness to share data for privacy reasons. Therefore, there is a fundamental need for designing distributed algorithms to perform the classical machine learning tasks.

There are major concerns one should consider before distributing their algorithm. First of all, we would like to get solutions with high accuracy as close as possible to what we get the classical algorithms on one machine. Second, communication between the machines would be often times the most time-consuming part of these algorithms. There are usually limitations in terms of the bandwidths and the amount of data we can transfer. In addition to the bandwidth issues, since the communication takes more time, our computing machines stay idle for a long time if we have too much communication. Hence, we would like to design algorithms with minimum or limited data transfer. Finally, one would expect to get faster solutions when multiple machines are available for the processing compared to a single machine. In an ideal case, when $m$ machines are available we wish the processing time by the factor of $m$. However, considering the time for communication this is not feasible. Instead, the existing solutions only try to get closer to this number.

\subsection{Problem formulation and framework}
Most supervised machine learning problems can be formulated as a stochastic optimization problem with the goal of minimizing an objective which is usually the expected loss. This can be written as:
\begin{equation}
\min_{w \in W} F(w) = \mathrm{E}_{z\sim \mathcal{D}} [f(w,z)]
\label{objective-equation}
\end{equation}
where $f$ is a given function (loss on one data point), $w$ is the unknown parameter vector and i.i.d samples $z_1, z_2, \ldots z_N$ are drawn from an unknown distribution $\mathcal{D}$. The problem to solve is to find the parameter vector $w$ which minimizes this expected loss.

In our setting we would assume that each machine $i, i \in {1, \ldots , m}$ has access to $n$ i.i.d samples $z_i^1, z_i^2, \ldots z_i^n$ from $\mathcal{D}$. Therefore, each machine can empirically construct its own local estimation of $F(w)$ as:
\begin{equation}
\phi_i(w) = \hat{F}_i(w) = \frac{1}{n}\sum_{j=1}^n f(w,z_i^j)
\end{equation}
and eventually the overall empirical estimation of $F$ is:
\begin{equation}
\label{empirical-loss}
\phi(w) = \hat{F}(w) =  \frac{1}{m}\sum_{i=1}^m \hat{F}_i(w,z_i^j) = \frac{1}{nm} \sum_{i, j} f(w, z_i^j)
\end{equation}
and we can minimize this empirical estimation to approximate the minimizer of $F$ in \ref{objective-equation} .

We focus our work on $L$-smooth and $\lambda$-strongly convex functions with $\kappa = L/\lambda$ being the condition number. A twice differentiable function $f$ is $L$-smooth and $\lambda$-strongly convex iff for all $w$ the Hessian of $f$ at $w$ is bounded by $\lambda$ and $L$, $\lambda \preceq \bigtriangledown^2 f(w) \preceq L$

In this work we are interested in central and synchronized approaches for solving this problem. We assume we have a central machine which communicates with the other computing machine. Since the individual data instances can be only accessed on the local machines, the algorithms consist of a series of rounds of local computations followed by some communication among the machines.
Therefore there would be some computation overhead for synchronous algorithms. Because we cannot perform the update until every node finishes their computation which is inefficient for the computation power of the nodes which finish their part faster. On the other hand since analyzing asynchronous solutions is difficult [\cite{federated}] thus to understand the solutions better we only consider synchronous approach in this work.

\subsection{Outline of our approach}
In this thesis we focus on DANE, a communication efficient algorithm for distributed optimization and try to improve it in terms of the amount of local computation which directly controls the overall run-time.
Will study different approximation algorithms for solving the sub-problems and replace the exact solver with the approximates algorithms. The algorithms will behave differently because of these approximate solvers. We find a right balance which is computationally less expensive but at the same time does not compensate too much accuracy in the final solution.

First, we study the most straightforward approximation algorithm, SGD and use it for a limited number of iterations for solving the local sub-problem inside DANE. In order to get speed-up in the convergence of SGD we use a variance reduced version, SVRG from \cite{svrg} and compare the two approaches and other distributed learning algorithms. After studying the problem in the batch mode where we assume using all of the available data we briefly discuss both the scenarios where we either can only use limited amount of data on the local machines in the very large dataset scenarios and also the streaming scenarios where we collect data on the go and need to find a suitable strategy to use the recent ones but gain benefits from the data we saw before. The remaining work on these two direction stays for the future work.

In section \ref{litret} we will review the literature on the optimization algorithms for this problem in the sequential and distributed settings. In section \ref{basics} we will introduce algorithms that we will use as the baselines in the experiments. In section \ref{methods} we propose our algorithms for approximate DANE. In section 5 we will discuss the need for the methods which only rely on the partial amount of data either because of the computational limitations or in applications where we have streaming data. Finally in section 6 we provide experimental results from the the approaches discussed and conclude with practical solutions. The last section concludes the paper and discusses ideas for future direction.

\section{Literature Review}
\label{litret}
\subsection{Stochastic optimization and learning}
A basic optimization algorithm which can be used to minimize the empirical loss in \ref{empirical-loss} is Gradient Descent (GD). At each step a descent direction is obtained by calculating the gradient of the function, from all samples, and used for finding the new iterate along with the descent direction. The algorithm is guaranteed provide an $\epsilon$-accurate solution in $O(\kappa \log \frac{1}{\epsilon})$ steps for strongly convex and smooth functions [\cite{nesterovbook}] and by using the accelerated version [\cite{nesterov}] we can get a better convergence rate of $O(\sqrt{\kappa} \log \frac{1}{\epsilon})$. When dealing with large amount of data, the main problem with using the gradient descent is the fact that the gradient is computed using all data at each iteration and this is not feasible in that settings.

As an alternative to gradient descent some stochastic versions of gradient descent are commonly used for applications with large sample size. These algorithms are almost the same as GD but the gradient is calculated using only one (SGD) or a constant number of samples (mini-batch SGD) instead of all samples. In these algorithms the stochastic gradient is still equal to the function's gradient in expectation. For strongly convex and smooth functions SGD converges in $O(\frac{L^2}{\lambda \epsilon} )$ iterations. This rate is worse in terms of the dependence on $\epsilon$ but the gain is mostly obtained by calculating the gradient only for one sample rather than $N$ of them at each iteration. In order to gain higher accuracy SGD is not very practical and choosing the right parameter is also not very straightforward.

%For machine learning problems the strong convexity often becomes available because of the regularization term. The regularization factor used is usually some $\lambda \sim |frac{1}{\sqrt{nm}}$ and given that, the number of iterations in both GD and SGD grow polynomially with the number of samples. 

The main reason why SGD converges slower than GD is the variance in the stochastic gradient which is not vanishing to zero. As a result, we need to use decaying step-size $\alpha_t = O(1/t)$. Using a decreasing step-size prevents diverging or zig-zagging around the optimal solution. As a consequence of this choice, SGD reaches moderate accuracy very fast but to achieve higher accuracies it needs too many iterations. To fix this problem, several approaches are  proposed to reduce the variance and provide better convergence results such as SVRG [\cite{svrg}] , SAG [\cite{sag}] and SAGA [\cite{saga}]. These algorithms provide similar convergence rates but SAG and SAGA rely on the storage of the gradients in a table and using them for the future calculations which is impractical in may applications. On the other SVRG does not rely on an extra storage, and the main cost it adds to SGD is calculating a full gradient using all samples occasionally. Using proper parameters, SVRG can achieve an accuracy of $\epsilon$ in $O(\log (1/\epsilon))$ stages and processing $O(n)$ gradients at each stage.

Another algorithm which can be considered as a variance reduction solution for SGD is SDCA [\cite{sdca}], Stochastic Dual Coordinate Ascent which is randomized coordinate ascent on the dual objective. SDCA solves the dual problem by choosing a dual coordinate at random at each iteration and optimizing with respect to that coordinate. Theoretical analysis link the guarantees for the dual solution to the primal and SCDA enjoys similar convergence properties as the approaches mentioned above and it is very popular in practice since it does not need choosing a learning rate. In [\cite{svrg}] a variance reduced view of SDCA is presented to make a connection between SDCA and SAG or SVRG as all being variance reduced methods for SGD. Similar to SAG and SAGA, SDCA also needs storage of the values as in SAG and hence is not suitable for all applications.

\subsection{Distributed optimization and learning}
%\section{Distributed Algorithms}
%one paragraph about the other algorithms:

When it comes to distributed optimization, one handy solution is to solve the problem separately on the machines and then take their average as the overall solution. However, it is shown in the previous studies [\cite{shamir}] that the simplest one-shot averaging approach provides an approximate minimizer
of $\phi(w)$ with finite suboptimality and we are not guaranteed to get any desired $\epsilon$-suboptimality. In practice the result can be much worse than the actual minimizer $w^*$ for Eq. \ref{objective-equation}.

Another alternatives is the gradient descent based solutions. Since with gradient descent each round of local computation is very expensive it is not practical in big data paradigm and thus we focus on the stochastic solutions. Based on the amount of the communication we allowed among the machines, we could have different variants of stochastic gradient descent (SGD). As one extreme, we can take the average of the local (stochastic) gradients on different machines at each step and use their average for the SGD iteration. The main problem with this solution is the excessive amount of communication as they send out their local gradient for all iterations. This builds up to huge amount of communication since the number of iterations needed for convergence of SGD is linear in the condition number $\kappa$ of the problem (condition number of the unknown underlying matrix) [\cite{shamir}]. In many machine learning problems, with considering the regularizer parameter $\lambda$ which is often of the order of $\sqrt{N}$, this leads to increase of the number of iterations as a polynomial in the number of total sample points $N$. In the big-data setting this would be highly impractical both in terms of communication and computation.

Another popular algorithm for distributed optimization is the alternation direction method of multipliers (ADMM) [\cite{dist-admm1, dist-admm2}]. ADMM takes the form of a decomposition-coordination procedure which uses dual decomposition and coordinated the local solutions to find a global solution to the problem. For strongly convex and smooth functions this approach can achieve linear convergence with complexity of $O(\sqrt{\kappa}log{(1/\epsilon)})$. For machine learning applications this means the increase in the number of iterations polynomially with $N$ which is not practical. There are other distributed algorithms, but mostly involving high communication cost since they communicate at each iteration and we do not get better iteration complexity in terms of their dependence on $\kappa$. Therefore it is reasonable to perform more computations locally on each machine before communicating with the other machines considering the imbalance between the communication and computation costs.

The above shortcomings of the first order methods has led researchers to develop second-order methods. A distributed approximate Newton-type method (DANE) is proposed in [\cite{dane}] which performs significant amount of computation at each iteration to solve a complete optimization sub-problem and find an exact solution to it. For quadratic functions DANE converges in $O(\frac{\kappa}{N}  log{(1/\epsilon)})$ with high probability [\cite{dane}]. Another second-order approach is DiSCO [\cite{disco}] which uses an inexact damped Newton method. Both DANE and DiSCO are very efficient with respect to their communication and are specially developed to fulfill this requirement. However, they both need extensive computation for each iteration which makes their running-time quite big.

On the other hand and in contrast to DANE and DiSCO, which find solutions for the sub-problems with a high accuracy and hence has high computational cost, COCOA and COCOA+ [\cite{cocoa, cocoa+}]  present a framework for a dual method which allows the use of local solvers with weak local approximation quality. With respect to the suboptimality of the primal function, COCOA+ quickly converges to modest accuracy, but for higher accuracies it converges very slowly \cite{federated}.

%The CoCoA framework formulates general local subproblems based on the dual form of (1) (See for instance [57, Eq. (2)]). Data points are distributed to nodes, along with corresponding
%13
%dual variables. Arbitrary optimization algorithm is used to attain a relative ? accuracy on the local subproblem ? by changing only local dual variables. These updates have their corresponding updates to primal variable w, which are synchronously aggregated (could be averaging, adding up, or anything in between; depending on the local subproblem formulation).
%From the description in this section it appears that the CoCoA framework is the only usable tool for the setting of Federated Optimization. However, the theoretical bound on number of rounds of communications for ill-conditioned problems scales with the number of nodes K. Indeed, as we will show in Section 4 on real data, CoCoA framework does converge very slowly.

%
Different versions of distributed SVRG are recently proposed. A variant of distributed SVRG is proposed in [\cite{svrg-iid}] which only uses the information of all machines to calculates the full gradients but afterwards performs the inner steps of SVRG sequentially on one local machine at a time. In order to maintain the SVRG convergence guarantees, they replicate a separate set of data and distribute to the machines to simulate i.i.d. sampling in 
distributed environment. Such setting is often time not practical in many applications. In [\cite{federated}], the authors study the problem in a rather different setting of federated optimization which is far from what we have here, with data which is not evenly distributed and is not an i.i.d sample of the main distribution on each machine. Their proposed algorithm is in its simple form a distributed version of SVRG that we will introduce in section \ref{methods} and extend it to suit the federated optimization setting. They also point to the equivalence of this algorithm to an approximation of DANE that we study here but without providing theoretical analysis or experimental studies on the plain version.

%A relatively similar method to Algorithm 3 presented here has been proposed in [78], which was analysed, and in [59], a largely experimental work that can be also cast as communication e�cient ? described in detail in Section 2.3.3.
%

% TODO: add a paragraph for aide

\section{Benchmark Algorithms}
\label{basics}
We consider some benchmark algorithms to compare our algorithms with. We will focus on showing our solutions can beat the first order algorithms while being being much more efficient in terms of the communication cost. Here we mostly focus on Stochastic Gradient Descent, SGD and Distributed Stochastic Gradient Descent, Dist-SGD as explained in the following sections.

\subsection{Stochastic gradient descent}
In big data scenarios, where the number of samples $N$ is very large, it is not practical to use Gradient Descent to solve the problems because of the high cost of calculations needed for calculating the gradient at each step. An alternative solution to this problem is Stochastic Gradient Descent or SGD which takes only one sample for the update at each iteration. Using only the gradient of one sample at each iteration makes the calculations for each SGD iteration very cheap as opposed to the Gradient Descent iterations.

Assume having  $z_1, z_2, \ldots z_N$ sampled i.i.d from distribution $\mathcal{D}$ and a function $F(w) =  \frac{1}{N} \sum_{i=1}^N f(w, z_i)$. In order to solve $\min_{w \in W} F(w)$, SGD starts with some initial iterate $w^0$, and continues by picking one stochastic sample $z_k$ at a time and using the gradient on this sample as the the update direction. The update formula would be of the form
\begin{equation}
w^{(k+1)} = w^k - \alpha_k (\bigtriangledown f(w^k , z_{k}))
\end{equation}
With step-size $\alpha_k$ for the $k$'th sample. The expectation of the stochastic gradients is the same as the gradient of the function. With this condition we can hope that the algorithm also behaves like gradient descent and there are theoretical guarantees for the convergence of SGD for strongly convex functions.

Although the computation at each iteration of SGD is much cheaper than gradient descent, the variance of the update term can typically be big which forces us to use a decreasing sequence of step-sizes $\{\alpha_k\}$. The convergence speed of SGD is therefore limited by these noisy approximations of the gradient. As a result for higher accuracies SGD takes a lot more iterations than the gradient descent. The theoretical analysis show the number of iterations needed to obtain $\epsilon$ accuracy is of the order of $O(\log 1/(\epsilon))$ for gradient descent and $O(1/\epsilon)$ for SGD which is by a logarithmic factor worse. Still when we have very large amount of data the total computation of gradient descent is $O(Nd \log 1/ \epsilon)$ which is higher than SGD's computation which is $O(d /\epsilon)$ and SGD is a better option.

\subsubsubsection{\bf{Setting the learning rate for SGD: }}
Choosing a proper step-size for SGD can be difficult. A step-size that is too small leads to very slow convergence, while large step-sizes can risk the convergence of the algorithm and cause the loss function to fluctuate around the minimum or even to diverge.

In order to set the step-sizes $\{\alpha_k\}$ we refer to the analysis used to provide convergence guarantees for SGD \cite{sgdhazan, sgdnemirovski, sgdrakhlin}. The convergence is obtained if $\sum_k {\alpha_k} = \infty$ and $\sum_k {\alpha_k} < \infty$ and using $\alpha_k = 1/(k+1)$ satisfies this condition. However, the best convergence speed is reached by using $\alpha_k = a_0/(1+\lambda k)$ where $a_0$ is the initial step-size and $\lambda$ specifies the step-size decay rate which is in the order of $1/k$. In our experiments we use this form for the step-size and choose the parameters which work better for our problem.

\subsubsection{Ideal distributed stochastic gradient descent}

In order to provide another benchmark for our experiments we assumed an ideal version of distributed SGD. What we mean by ideal is the case where distributing the SGD algorithm does not introduce any additional costs, thus the computation is only divided by the number of the machines $m$. We mainly use this distributed notion in our experiments to show how the other algorithms stand compared to this imaginary algorithm. We add it to our experiments by simply running SGD on only one machine and dividing the number of the gradients calculated by $m$. You will see this curve used in many plots in the experiments section.

\subsection{Distributed stochastic gradient descent}
In this section we will introduce a distributed implementation of SGD that we use in our experiments. The main idea would be to have the machines run SGD on their own and once in a while communicate with each other. Considering the amount of communication, we can consider two extremes. One would be to having the machines one step of SGD and communicate either the gradient or the iterate with each other and use the average for the next iteration. This version would essentially be a parallel implementation of Mini-Batch Gradient Descent where we take the mini-batch size $B$ equal to $m$. On the other extreme we can have all the machines run a full SGD for enough iterations that is needed and only communicate their final iterates and take their average as the solution \cite{par-sgd}.

As it is clear there is a trade-off between the number of communication rounds and the amount of computation at each round on the local machines. As we increase the local computation we need lower number of communication rounds. At the same time since communication needs a lot more time, by communicating in too many rounds in addition to increasing the communication cost which is expensive we also increase the idle time on the local communication machines which is not desirable for the overall efficiency of our algorithm.

Here, we take the algorithm for the general case, where we use $T$ to specify the number of SGD steps in each machine before communicating with the others. You can see this approach in algorithm \ref{dist-sgd}. Choosing $T=1$ leaves us with the first extreme discussed above while choosing only one round of communication and running local SGD algorithms as much as needed leaves us with the second extreme. For our experiments we mostly use the same amount of local computations which is aligned with the other algorithms that we use.

\begin{algorithm}[t]  
\caption{{\bf Distributed SGD}}
\label{dist-sgd}
\begin{algorithmic}[1]
\State {\bf Parameter:} T = number of SGD steps per communication round
\State {\bf Initialize:} Start at some $w^{(0)}$
\State {\bf Iterate:}
\For t = 1, 2, \ldots  
\For{each machine $i$}
\State $w_i = w^{(t-1)}$
\For{$k = 1, 2, \ldots , T$ } 
\State pick $z_k$ in random from $\{1, 2, \ldots , n\}$
\State $w_i = w_i - \alpha_k \bigtriangledown f(w_i , z_k)$
\EndFor
\EndFor
\State Compute $w^{(t)} = \frac{1}{m} \sum_{i=1}^m w_i$ and distribute to all machines
\EndFor
\end{algorithmic}
\end{algorithm}

\subsubsubsection{\bf{Setting the learning rate for distributed SGD: }}
In order to set the step-sizes $\{\alpha_k\}$ for Distributed SGD we again use the same form as we explained for SGD which is using $\alpha_k = a_0/(1+\lambda k) $. We will choose parameters suitable for this algorithm.

% section 4:
\section{DANE with Approximate Local Solvers} \label{methods}

In this section we would first explain  Distributed Approximate NEwton or DANE, the main algorithm that we study in this paper and explain why although it satisfies the important condition of having low communication loss it is not very practical because of having extensive local computations. Then we would proceed by providing some alternative approximate approaches for its local calculations and study how they change the behavior of the algorithm.

\subsection{DANE: Distributed Approximate NEwton}

Distributed Approximate NEwton, DANE, is an approximate
Newton-like method \cite{dane}. This method uses descent steps
which are chosen based on the geometry of the problem at each iteration. The main benefit over the first order approaches is using more appropriate steps and hence decreasing the number of iterations needed to obtain the desirable accuracy.

This algorithm uses a central machine and consists of the following main steps which are performed in a sequence. It calculates the current iterate $w^t$ and gradient $\bigtriangledown \phi(w^t)$ by collecting the iterates and local gradients from the machines and taking their average and distributes these values back to the machines. In the second step all of the computing machines solve an optimization sub-problem and send back the solution to the central machine.

Instead of minimizing the original function $\phi_i(w)$ on the local machines, which would lead to have the average of the optimal solutions as the final result, DANE forms an objective function by mainly adding a perturbation term in order to take into account the second-order information. You can see DANE altogether in algorithm \ref{DANE}. The local subproblem is
\begin{equation}
w_i^{(t)} = \argmin_w [\phi_i(w)- (\bigtriangledown \phi_i(w^{(t-1)}) - \eta  \bigtriangledown \phi(w^{(t-1)}))^T w + \frac{\mu}{2} ||w-w^{(t-1)}||^2_2]
\end{equation}
where $\bigtriangledown \phi_i(w^{(t-1)})$ is the gradient on machine $i$ and $\bigtriangledown \phi(w^{(t-1)})$ is the global gradient which is the average of the local gradients, both at the previous iterate $w^{(t-1)}$.

For a quadratic function this subproblem leads to an approximation of the Newton update. The analysis of this algorithm \cite{dane} are also provided for the quadratic functions. However, we could consider the quadratic approximation of the functions in small neighborhoods and expect to observe similar behavior when we are close enough to the optimal solution. In practice DANE performs quite well on non-quadratic functions a well.

%TODO: --------  needs more explanation of the local problem and intuition
%--------  needs more explanation of the quadratic case
%--------  needs discussing a bit the guarantees provided in the paper ( -------- discuss the solutions and their complexity )
%

\begin{algorithm}[t]
\caption{{\bf DANE}  \cite{dane}}
\label{DANE}
\begin{algorithmic}[1]
\State {\bf Parameter:} learning rate $\eta > 0$ and regularizer $\mu > 0$
\State {\bf Initialize:} Start at some $w^{(0)}$, e.g. $w^{(0)} = 0$
\State {\bf Iterate:}
\For{for t = 1, 2, \ldots  }
\State Compute $\bigtriangledown \phi(w^{(t-1)}) = \frac{1}{m} \sum_{i=1}^{m} \bigtriangledown  \phi_i(w^{(t-1)})$ and distribute to all machines
\State For each machine $i$, solve

$w_i^{(t)} = \argmin_w [\phi_i(w)- (\bigtriangledown \phi_i(w^{(t-1)}) - \eta  \bigtriangledown \phi(w^{(t-1)}))^T w + \frac{\mu}{2} ||w-w^{(t-1)}||^2_2]$
\State Compute $w^{(t)} = \frac{1}{m} \sum_{i=1}^m w_i^{(t)}$ and distribute to all machines
\EndFor
\end{algorithmic}
\end{algorithm}

DANE converges in a very few number of iterations which means also few round of communication among the machines. It also only communicates first order information (gradients) which leaves us with a very low communication cost. On the other hand each machine finds an exact solution to the local optimization problem in line 6 of the algorithm which is computationally expensive. This problems limits the applications of DANE in solving the distributed problems. It will not even able to compete SGD if we consider their computation cost. In order to address this problem we will combine stochastic optimization with DANE in the following sections to avoid the exact calculations in the machines and obtain a computationally efficient solution.

\subsection{DANE combined with approximate local solvers}
One solution for reducing the cost of using exact local solvers in DANE is to replace them with stochastic solutions. The first option which bears in mind is Stochastic Gradient Descent (SGD) which is the stochastic alternative to Gradient Descent and is much more applicable to the big data regime. In the following subsections we provide solutions based on using SGD and SVRG, which is a reduced variance alternative to SGD, for the local solvers. This would lead to Algorithm xxx.

\subsubsection{DANE combined with SGD}

We would like to relax the DANE procedure my replacing the exact solver by approximate SGD updates for the local solvers. For each DANE iteration $t$ we can initialize $w_i = w^{t-1}$ for each machine $i$ and apply the updates for solving 
\begin{equation}
w_i^{(t)} = \argmin_w [\phi_i(w)- (\bigtriangledown \phi_i(w^{(t-1)}) - \eta  \bigtriangledown \phi(w^{(t-1)}))^T w + \frac{\mu}{2} ||w-w^{(t-1)}||^2_2]
\end{equation}
Taking the learning rate $\alpha $ (we will discuss it later) the SGD updates are:
\begin{equation}
w_i = w_i - \alpha (\bigtriangledown f(w_i , z_k)- ( \bigtriangledown \phi_i(w^{(t-1)}) - \eta  \bigtriangledown \phi(w^{(t-1)})) + \mu(w-w^{(t-1)}))
\label{danesgd-update}
\end{equation}
After applying our desired steps of SGD we set $w_i^t = w_i$ and proceed with the rest of the DANE procedure as explained in algorithm \ref{DANEsgd}. 

\begin{algorithm}[t]
\caption{{\bf DANE combined with SGD}}
\label{DANEsgd}
\begin{algorithmic}[1]
\State {\bf Parameter:} learning rate $\eta > 0$, regularizer $\mu > 0$, T = number of SGD steps per DANE iteration
\State {\bf Initialize:} Start at some $w^{(0)}$, e.g. $w^{(0)} = 0$
\State {\bf Iterate:}
\For t = 1, 2, \ldots  
\State Compute $G = \bigtriangledown \phi(w^{(t-1)}) = \frac{1}{m} \sum_{i=1}^{m} \bigtriangledown  \phi_i(w^{(t-1)})$ and distribute to machines
\For{each machine $i$}
\State $w_i = w^{(t-1)}$
\State $G_i = \bigtriangledown \phi_i(w^{(t-1)})$
\For{$k = 1, 2, \ldots , T$ } 
\State pick $z_k$ in random from $\{1, 2, \ldots , n\}$

\State $w_i = w_i - \alpha (\bigtriangledown f(w_i , z_k)-  G_i + \eta G + \mu(w-w^{(t-1)}) )$
\EndFor
\EndFor
\State Compute $w^{(t)} = \frac{1}{m} \sum_{i=1}^m w_i$ and distribute to all machines
\EndFor
\end{algorithmic}
\end{algorithm}

\subsubsubsection{\bf{Setting the learning rate for SGD: }}
From the convergence analysis of SGD we know that it is not guaranteed to converge with a fixed learning rate. We also know using a suitable step size usually affects the SGD convergence behavior to a great extent and is very important. We typically need to use learning rates in the form of  $\alpha_k = a_0/(1+\lambda k)$ to get the best results. However, it is not strange that this formula is not necessarily useful for the combination of DANE and SGD. Because DANE converged a lot faster than SGD, at each iteration $w^t$ gets exponentially closer to the optimal $w$ if we have exact DANE and therefore, the step size should also get smaller appropriate to this rate. For our experiments we set the step size to be of the form $\alpha_k =a_0 / (\exp{(c t)} (1+\lambda k) )$ where $t$ is the DANE iteration, $k$ is the SGD iteration number and $c$ is a constant.

%TODO ----------- continue this part -  refer both to the DANE convergence analysis and also the SGD convergence analysis to explain and justify this part.

%\subsection{DANE with mini-batch SGD local solver}
\subsection{DANE combined with SVRG}
DANE combined with SGD as the local solver performs quite well if we do not seek very high precision results. However, because of the decaying learning rate that we should use in SGD in order to have convergence it takes a lot of iterations to get to more exact solutions for the local subproblems. To avoid this problem one can use variance reduced alternatives for SGD \cite{svrg, sag, saga} which converge with a fixed learning rate normally bigger than the rate used in the later iterations in SGD. One of these algorithms is SVRG which uses a different descent direction that is still equal to the function gradient in expectation but has smaller variance that converges to zero as we get closer to the optimal solution. 

The SVRG algorithm consists of several stages. At each stage a series of SVRG updates are performed. Each update integrates an approximate iterate $\tilde{w}$ and approximate full gradient $h$ at $\tilde{w}$ specific to that stage. The value of $\tilde{w}$ is obtained from the previous stage either by taking the last iterate, or the average of the last iterates. The value of $h$ is calculated as the average of the gradient of all samples which needs a full pass over the dataset. For an arbitrary function of the form
\begin{equation}
F^{*}(w)  = \frac{1}{N}\sum_{i=1}^N f^{*}(w,z_i)
\label{Fstar}
\end{equation}
the descent direction calculated using sample $z_k$ (which we denote by $g_k$) is:
\begin{equation}
\begin{aligned}
g_k = \bigtriangledown f^{*}(w , z_k)- \bigtriangledown f^{*}(\tilde{w} , z_k) + \bigtriangledown F^{*}(\tilde{w}) \\
= \bigtriangledown f^{*}(w , z_k)- \bigtriangledown f^{*}(\tilde{w} , z_k) + h 
\end{aligned}
\label{svrgdescent}
\end{equation}
and the iterate updates will be of the form
\begin{equation}
w = w - \alpha( \bigtriangledown f^{*}(w , z_k)- \bigtriangledown f^{*}(\tilde{w} , z_k) + h )
\end{equation}
where $\alpha$ is the learning rate and could be a fixed value.

In the DANE algorithm we already have calculated an average iterate $w^{(t-1)}$ and full gradient $G = \bigtriangledown \phi(w^{(t-1)}) $ before each DANE iteration $t$. One way to apply SVRG on the DANE sub-problems without additional costs of calculating more full gradient for the SVRG stage is to use only one stage of SVRG for each iteration of DANE. This way we could simply use the average iterate $w^{(t-1)}$ and the full gradient $G$ as $\tilde{w}$ and $h$ for SVRG. For the rest of this paper we use single-stage SVRG as our local SVRG solver.

When using SVRG as the local solver we should notice that we want to solve the sub-problem on each computing machine $i$. In this case we would apply SVRG on
\begin{equation}
F^{*}(w) = \phi_i(w)- (\bigtriangledown \phi_i(w^{(t-1)}) - \eta  \bigtriangledown \phi(w^{(t-1)}))^T w + \frac{\mu}{2} ||w-w^{(t-1)}||^2_2
\end{equation}
which is different from  $\phi_i(w)$. In the above formula both $\bigtriangledown \phi(w^{(t-1)})$ and $\bigtriangledown \phi_i(w^{(t-1)})$ are fixed values for each DANE iteration and we denote them respectively by  $G$ and $G_i$. Substituting $G$, $G_i$ and $\tilde{w} = w^{(t-1)}$ (as we discussed above) into the formula we have:
\begin{equation}
F^{*}(w) = \phi_i(w)- (G_i - \eta  G )^T w + \frac{\mu}{2} ||w-\tilde{w}||^2_2
\end{equation}
and we can write $F^{*}(w)$ as $F^{*}(w)  = \frac{1}{N}\sum_{i=1}^N f^{*}(w,z_i)$ (Eq. \ref{Fstar}) with
\begin{equation}
\label{subproblem}
f^{*}(w , z_i) = f(w, z_i) - (G_i - \eta  G )^T w + \frac{\mu}{2} ||w-\tilde{w}||^2_2
\end{equation}

The DANE algorithm combined with using the SVRG update rules for the above function leads us to algorithm \ref{DANEsvrg}. We include the derivation of the update term in the following subsection \ref{svrgderiv}.

\begin{algorithm}[t]
\caption{{\bf DANE combined with SVRG} }
\label{DANEsvrg}
\begin{algorithmic}[1]
\State {\bf Parameter:} learning rate $\eta > 0$, regularizer $\mu > 0$, T = number of SVRG steps per DANE iteration
\State {\bf Initialize:} Start at some $w^{(0)}$, e.g. $w^{(0)} = 0$
\State {\bf Iterate:}
\For t = 1, 2, \ldots  
\State Compute $G = \bigtriangledown \phi(w^{(t-1)}) = \frac{1}{m} \sum_{i=1}^{m} \bigtriangledown  \phi_i(w^{(t-1)})$ and distribute to machines
\For{each machine $i$}
\State $w_i = w^{(t-1)}$
\For{$k = 1, 2, \ldots , T$ } 
\State pick $z_k$ in random from $\{1, 2, \ldots , n\}$

\State $w_i = w_i - \alpha (\bigtriangledown f(w_i , z_k)- \bigtriangledown f(w^{(t-1)} , z_k) + \eta G + \mu(w-w^{(t-1)}))$
\EndFor
\EndFor
\State Compute $w^{(t)} = \frac{1}{m} \sum_{i=1}^m w_i$ and distribute to all machines
\EndFor
\end{algorithmic}
\end{algorithm}

It is worth mentioning that we also experimented with different ways of using SVRG for solving the sub-problems. However we did not discuss using SVRG with more than one stage in this paper since in our experiments those solutions appeared to be considerably inferior to the current version.

\subsubsubsection{\bf{Setting the learning rate for SVRG: }}
Using SVRG has the benefit of keeping the variance of the update term small. Therefore it allows us to keep a fixed step-size rather than a decreasing one. As a result in order to set the step-size for SVRG updates we could set it by performing cross-validation over a set of parameters.

\subsection{Equivalence of DANE combined with SVRG to distributed SVRG}
\label{svrgderiv}

In this subsection we derive the formula for SVRG update rule when used to solve the sub-problems in DANE and show the equivalence of DANE combined with SRVG to a distributed version of SVRG on the main function.

Considering the SVRG as update rules for solving the sub-problem which is the average of the functions $f^{*}(w , z_i)$ (Eq. \ref{subproblem}) we can derive the SVRG descent direction $g_k$ introduced in \ref{svrgdescent} for this function as follows:
\begin{equation}
\begin{aligned}
g_k & = \bigtriangledown f^{*}(w , z_k)- \bigtriangledown f^{*}(\tilde{w} , z_k) + \bigtriangledown F^{*}(\tilde{w}) \\
& = \big{(}  \bigtriangledown f(w, z_i) - (G_i - \eta  G ) + \mu (w-\tilde{w})  \big{)} \\
& - \big{(} \bigtriangledown f(\tilde{w}, z_i) - (G_i - \eta  G )   \big{)} 
 + \big{(} \bigtriangledown \phi_i(\tilde{w})- (G_i - \eta  G )  \big{)} \\
& =   \bigtriangledown f(w, z_i) - \bigtriangledown f(\tilde{w}, z_i) + \bigtriangledown \phi_i(\tilde{w}) - (G_i - \eta  G ) + \mu (w-\tilde{w})  \\
\end{aligned}
\end{equation}

Noticing that $\tilde{w} = w^{t-1}$ we know $ \bigtriangledown \phi_i(\tilde{w})  = G_i$ and we get
\begin{equation}
g_k =   \bigtriangledown f(w, z_i) - \bigtriangledown f(\tilde{w}, z_i) + \eta  G + \mu (w-\tilde{w})  \\
\label{danesvrg-update}
\end{equation}

This $g_k$ is the update rule used in the inner loop in algorithm \ref{DANEsvrg}.

The update formula for the local solver in DANE combined with SVRG \ref{danesvrg-update} is in fact the same as the SVRG update formula if applied function $F$ itself. The only difference is the parameter $\eta$ which controls the effect of $G$ and the additional term $\mu (w-\tilde{w})$ in the DANE version. If we consider the most common setting of DANE which is using $\eta=1$ and $\mu=0$ the two algorithms are exactly equivalent. This result is also discovered in \cite{federated, aide}. This observation is very interesting since there are approaches for solving the distributed optimization problem both based on DANE and SVRG and this equivalence could explain some of the behaviors of these algorithms.

\subsubsection{Discussion on DANE combined with SGD versus SVRG}
In order to compare the two methods we first look at their update terms in the inner loop. Comparing the SVRG update formula \ref{danesvrg-update} to the update term of SGD as the local solver \ref{danesgd-update} we can notice there is a slight difference between the two. The term $\bigtriangledown\phi_i (w^{(t-1)})$ in the SGD case which is a fixed value calculated once for each DANE iteration using iterate $w^{(t-1)}$ and data on machine $i$ is replaced by the stochastic term $\bigtriangledown f (w^{(t-1)}, z_k)$ which would be again computed using iterate $w^{(t-1)}$ but this time only for data sample $z_k$.
This looks like a very small difference but in practice makes the variance in the SVRG case much smaller which provides faster convergence for getting higher accuracy.

Another aspect to consider is the fact that for SGD we only calculate the gradient $\bigtriangledown f (w_i, z_k)$ for the sampled datapoint $z_k$ at each local iteration, in contrast to SVRG where we have two calculations $\bigtriangledown f (w_i, z_k)$  and $\bigtriangledown f (w^{(t-1)}, z_k)$, and therefore the computational cost would be higher if we want to rum them for the same number of iterations. However, we could also use fewer iterations for SVRG to keep the two approaches comparable. We would provide results using these two different choices in the experiments section.

\section{Experiments}
\label{expr}

In this section we present our experimental results obtained from the algorithms we discussed in the previous sections. We explain our experimental set-up and the problem we try to solve as well as the results we obtain and the conclusions and insights from those experiments. For the most of the experiments we use the simple choice of setting $\eta = 1$ and $\mu = 0$. It is discussed in \cite{dane} that in some situations for instance where the amount of data per machine is small, the algorithm might not converge with these choices and we might benefit from using some positive value for $\mu$. In this case the convergence would be slower than when $\mu = 0$ is used.

\subsection{Experimental set-up}
{\bf Dataset:} We use the synthetic data in \cite{dane}. The training examples $(x, y)$ are sampled according to the model  $y = <x,w^*> + \xi$, $x \sim \mathcal{N}(0,\Sigma) $, $\xi \sim \mathcal{N}(0, 1)$ with $x, w^* \in \mathbb{R}^{500}$, the diagonal covariance matrix $\Sigma$ with $\Sigma_{i,i}i^{-1.2}$ and the parameter vector $w^*$ being an all-ones vector. We generate $N$ i.i.d samples from this model and distribute them randomly between the machines with even number of data points per machine. The problem we would like to solve here is the standard ridge regression problem $\min_w  \sum_{i=1}^N (<x, w> - y)^2 + 0.005 w^2$. In the following sections we use the algorithms discussed in the previous sections under different conditions to solve our ridge regression problem.

{\bf Evaluation measure:} In order to evaluate the algorithms we mainly consider the difference between the function's value on the algorithm's solution and the its value on the optimal parameter $w^*$. Measuring this value on the training data gives us the sub-optimality of the algorithm. We are also interested in the generalization of our solution and in order to measure that we should look into the population error for which we use a new set of data-points obtained from the same model and big enough to represent the population with high probability.

\subsection{Experiments with full data access}
In this section we look at the results where every data-point is locally accessible by it's own machine and the local computations can be performed using the whole data. This is not always possible for computational cost or delays or in streaming applications. We will bring some experiments for these scenarios in the following section. 
%- show training error and test error

\subsubsection{DANE combined with SGD and SVRG with different accuracies}

\subsubsubsection{Effect of the number of SGD steps }

\begin{figure}[ht]
\begin{subfigure}{.5\textwidth}
  \centering
  \includegraphics[width=1.\linewidth]{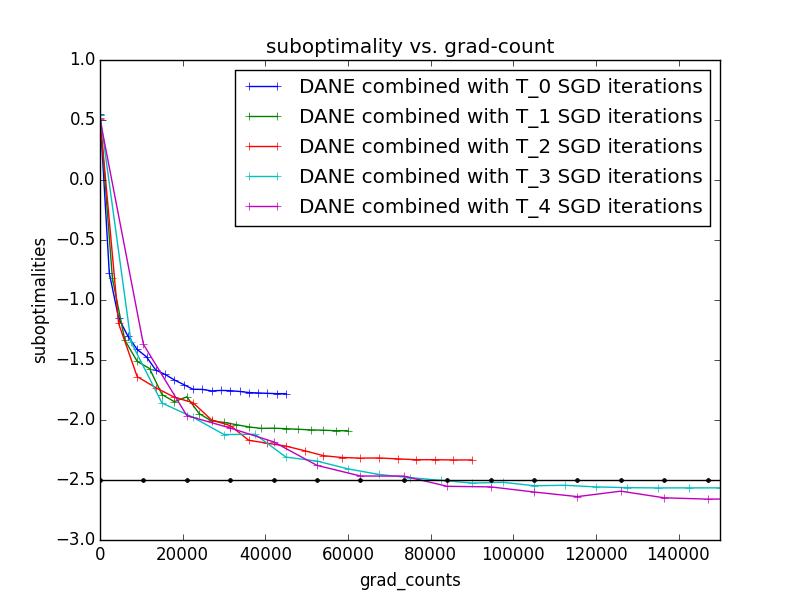}
  \caption{\scriptsize $(N , m) = (6000, 4)$ \normalsize}
  \label{fig:sfig1}
\end{subfigure}%
\begin{subfigure}{.5\textwidth}
  \centering
  \includegraphics[width=1.\linewidth]{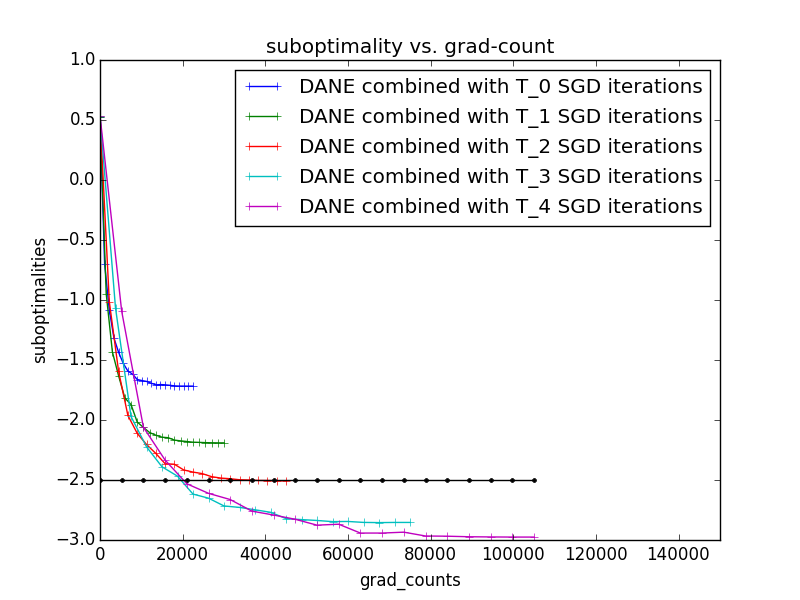}
  \caption{\scriptsize $(N , m) = (12000, 16)$ \normalsize}
  \label{fig:sfig2}
\end{subfigure} 
\begin{subfigure}{.5\textwidth}
  \centering
  \includegraphics[width=1.\linewidth]{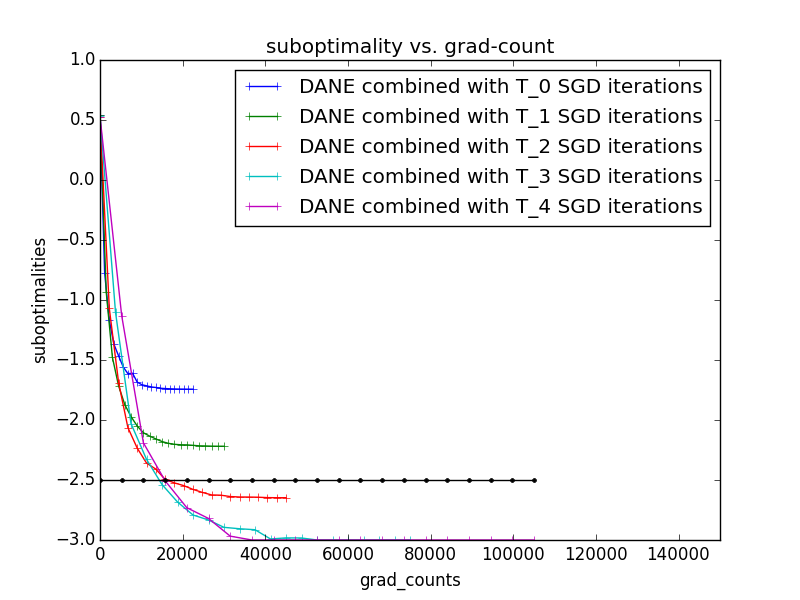}
  \caption{\scriptsize $(N , m) = (32000, 32)$ \normalsize}
  \label{fig:sfig2}
\end{subfigure}
\caption{DANE combined with SGD using different number of steps $ T \in \{0.5n, n, 2n, 4n, 6n \}$.}
  \label{fig:danesgd-different-T}
\end{figure}

We first make some experiments with DANE algorithm when it is combined with SGD. As mentioned earlier, in order to reach to very accurate solutions with SGD as the local solver many iterations are needed which is not our purpose here. However, in many learning problems lower accuracy is enough and using SGD could be a good option in those cases. Therefore we set our goal for this part to gain some given accuracy using DANE combined with SGD. We ran experiments with different numbers of stochastic gradient steps, $T$, that are taken inside each iteration of DANE on one machine. This helps us get an idea on how to choose the balance between the number of inner SGD steps compared to the number of DANE iterations. Figure \ref{fig:danesgd-different-T} shows the sub-optimality curves obtained from running the algorithms with different numbers of SGD steps $T$ from $\{T_0=0.5n, T_1=n, T_2=2n, T_3=4n, T_4=6n\}$ where $n$ is the number of samples on each machine and $T= xn$ indicates that the algorithm can visit and process as much as $x$ times the data on that local machine. We run the experiments with different set-ups in terms of the number of data-points and machines taking three different cases $(N , m) = \{(6000, 4), (12000, 16), (32000, 32)\}$. We specify the log-sub-optimality of $-2.5$ to compare different setting based on reaching to this accuracy. Since the most expensive part of the algorithms we have here is the calculation of the gradients, the x-axis shows the number of gradient calculations on each computing machine and the y-axis shows the sub-optimality reached by having those.

As it is evident from the figures, to reach to a fixed accuracy, it is better to use less SGD iterations as far as they are enough to reach us to the desired accuracy. If we do not take enough SGD iterations DANE will not contain its characteristics anymore and we might not get to that accuracy at all. Having more DANE iterations give us a better convergence rate from sole computational point of view, however, we should keep in mind that increasing the number of DANE iterations is equivalent to extra communication cost which should be balanced considering our system's specifications.

\subsubsubsection{Effect of the number of SVRG steps}

For DANE combined with SVRG the results are much more promising. Again we would like to have an idea on what is the best combination of the number of DANE iterations and SVRG inner steps to get to a given accuracy.

\begin{figure}[h]
\begin{subfigure}{.5\textwidth}
  \centering
  \includegraphics[width=1.\linewidth]{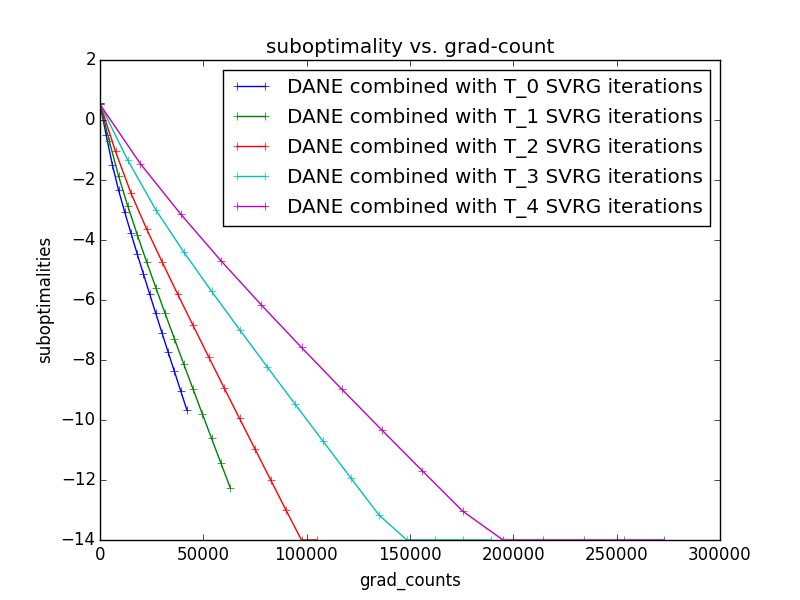}
  \caption{\scriptsize $(N , m) = (6000, 4)$ \normalsize}
  \label{fig:sfig1}
\end{subfigure}%
\begin{subfigure}{.5\textwidth}
  \centering
  \includegraphics[width=1.\linewidth]{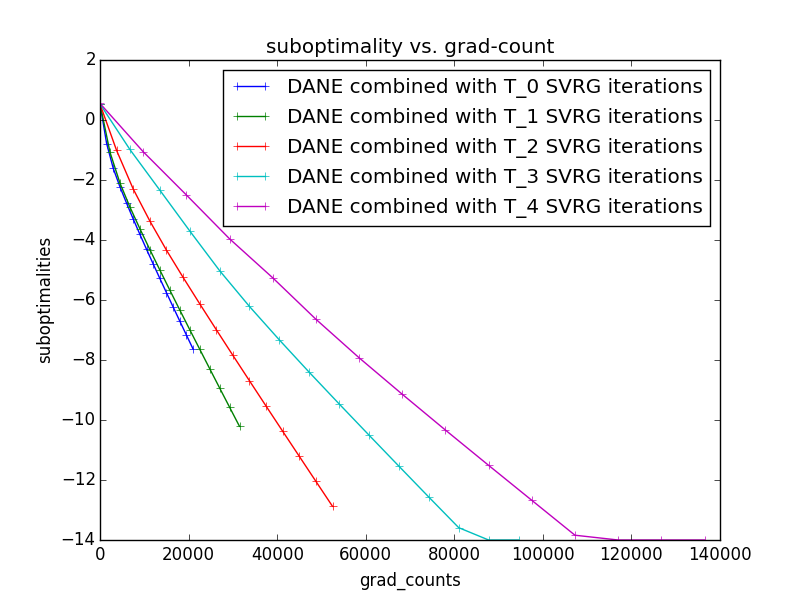}
  \caption\scriptsize {$(N , m) = (12000, 16)$ \normalsize}
  \label{fig:sfig2}
\end{subfigure} 
\begin{subfigure}{.5\textwidth}
  \centering
  \includegraphics[width=1.\linewidth]{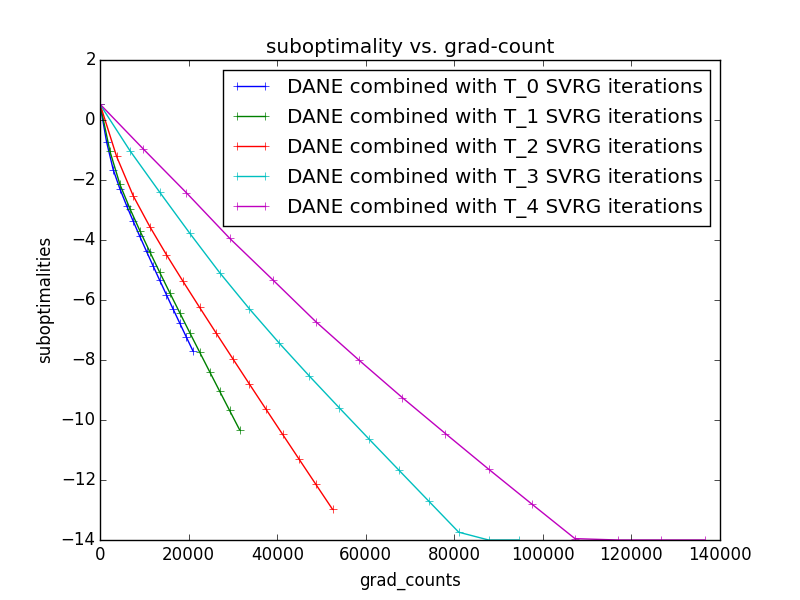}
  \caption{\scriptsize $(N , m) = (32000, 32)$  \normalsize}
  \label{fig:sfig2}
\end{subfigure}
\caption{DANE combined with SVRG, using different number of steps $ T \in \{0.5n, n, 2n, 4n, 6n \}$.}
  \label{fig:danesvrg-different-T}
\end{figure}

To see the effect of the local solutions accuracy, in Figure  \ref{fig:danesvrg-different-T} we illustrate the results of our experiments with different number of steps $T$ allowed for SVRG local solver and run experiments with the same setting as we had for DANE combined with SGD. Our algorithm shows linear convergence and again with more DANE iteration we get faster convergence to our desired accuracy. In contract to the experiments with SGD local solver, using lower number of SVRG inner steps does not prevent DANE to demonstrate the same behavior as in the cases where it enjoys more accurate solutions. With lower SVRG iterations, DANE still converges linearly to the optimal solution. However, the cost of communication is the rational for not using too many DANE iterations in contract to more steps of SVRG in the local solvers. Moreover, since the calculation of the global gradients needs one pass over the data it seems reasonable to keep the number of SVRG steps $T$ to be at least of the same order of $n$ for practical usages.

\begin{figure}[htbp]
\begin{subfigure}{.5\textwidth}
  \centering
  \includegraphics[width=1.\linewidth]{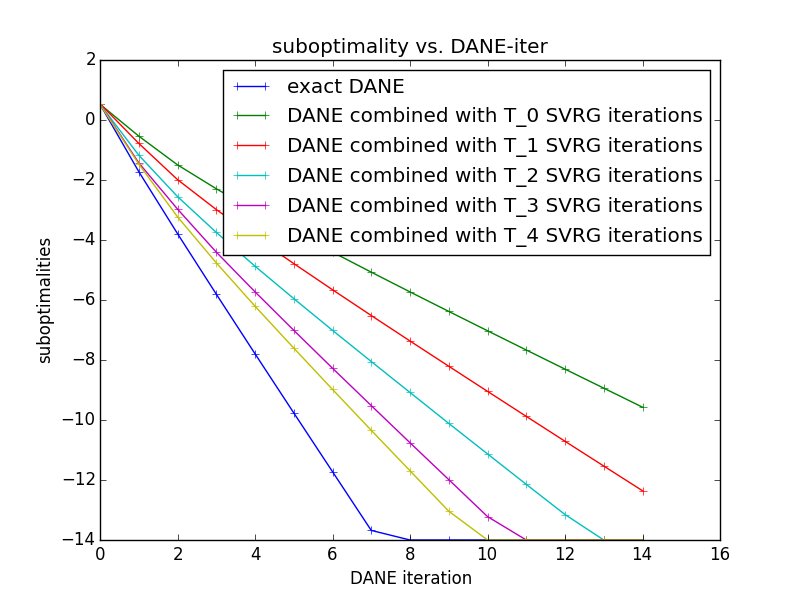}
  \caption{\scriptsize $(N , m) = (6000, 4)$ \normalsize }
\end{subfigure}%
\begin{subfigure}{.5\textwidth}
  \centering
  \includegraphics[width=1.\linewidth]{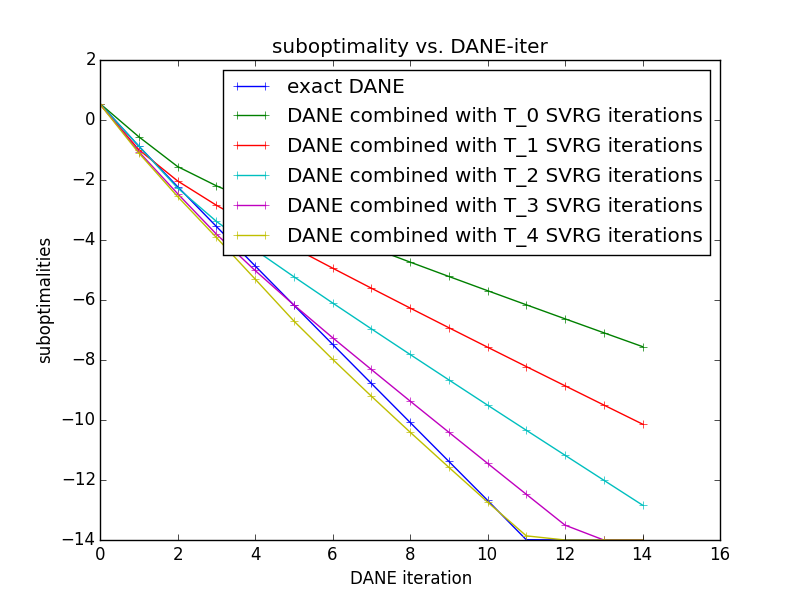}
    \caption{\scriptsize $(N , m) = (12000, 16)$ \normalsize }
\end{subfigure}%
\caption{Exact DANE vs. DANE combined with SVRG and $ T \in \{0.5n, n, 2n, 4n, 6n \}$.}
\label{fig:exact-vs-svrg}
\end{figure}

As it is evident from the results, DANE combined with SVRG performs very well. In order to really see how good this approximation works, in Figure \ref{fig:exact-vs-svrg} we illustrate the results from the DANE and compare it with DANE combined with SVRG using different number of SVRG steps per DANE iteration. Using enough number of SVRG iterations, makes the approximate DANE very close to the exact DANE and provides the exact same behavior. This is achievable with visiting about 5-6 epochs of data, but using less SVRG iterations will not have a very bad effect on this either. As you can see with using less stochastic steps we can still get to very high accuracies, however, more DANE iterations are needed for getting there. As we can see in Figure \ref{fig:danesvrg-different-T}, in general increasing the number of SVRG steps makes the algorithm's computation cost (total gradient calculations) higher, and it is only reasonable in order to reduce the amount of communication.

\subsubsection{Comparing approximate DANE to the baselines:}

\begin{figure}[htbp]
\begin{subfigure}{.5\textwidth}
  \centering
  \includegraphics[width=1.\linewidth]{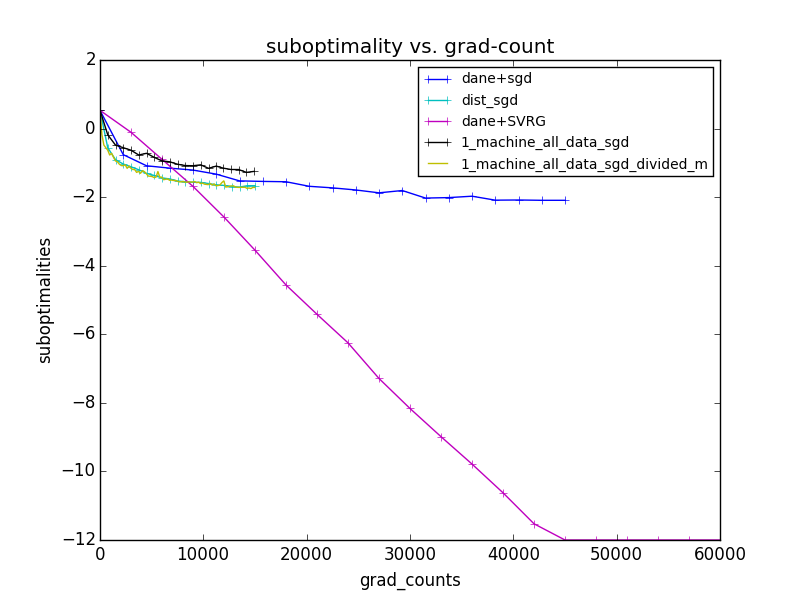}
  \caption{$(N , m) = (6000, 4), T = 0.5 n$}
  \label{fig:sfig1}
\end{subfigure}%
\begin{subfigure}{.5\textwidth}
  \centering
  \includegraphics[width=1.\linewidth]{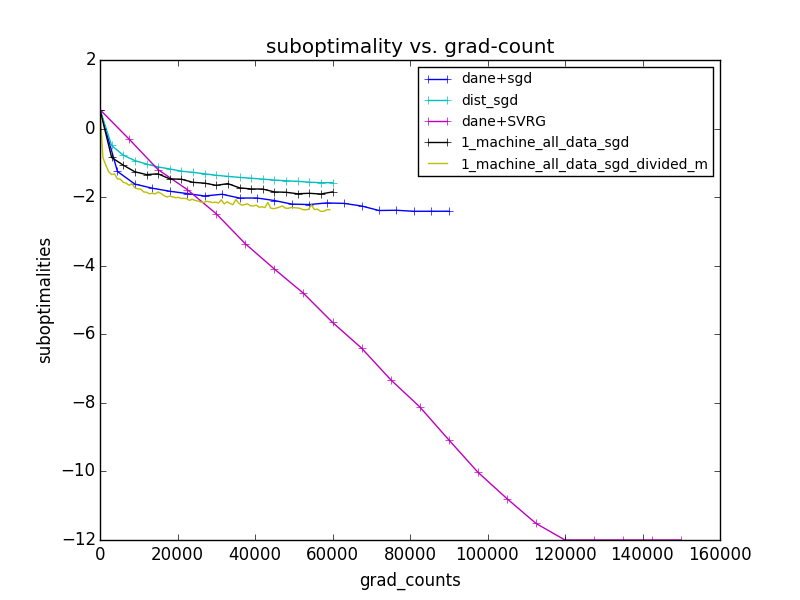}
  \caption{$(N , m) = (6000, 4), T = 2n$}
  \label{fig:sfig2}
\end{subfigure} 
\begin{subfigure}{.5\textwidth}
  \centering
  \includegraphics[width=1.\linewidth]{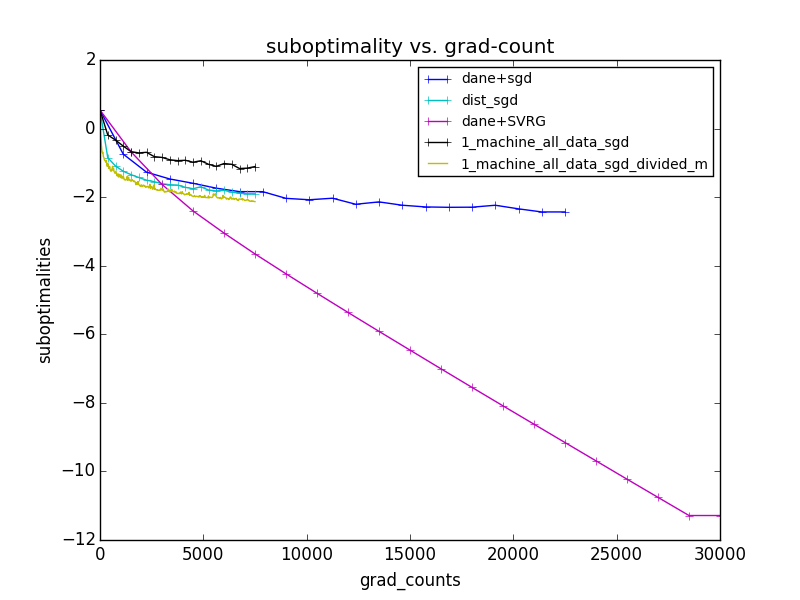}
  \caption{$(N , m) = (12000, 16),  T = 0.5n$}
  \label{fig:sfig2}
\end{subfigure}
\begin{subfigure}{.5\textwidth}
  \centering
  \includegraphics[width=1.\linewidth]{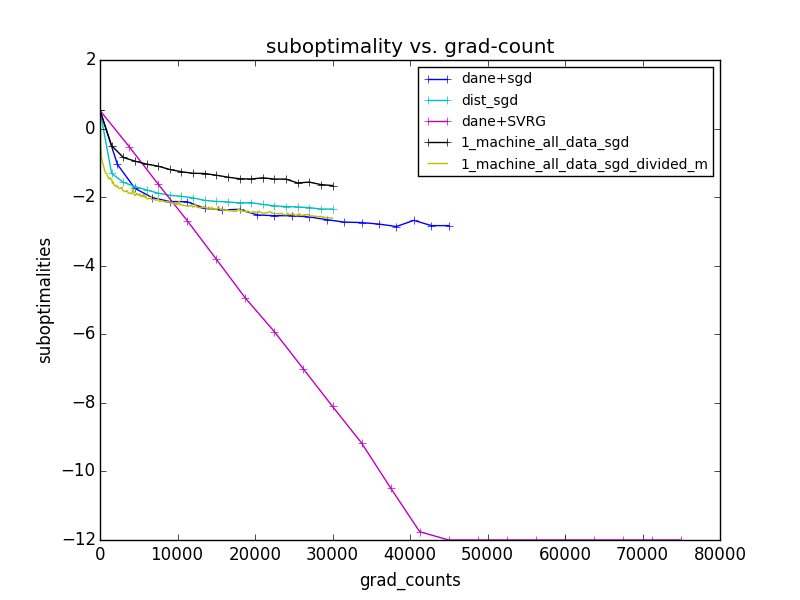}
  \caption{$(N , m) = (12000, 16), T = 2n$}
  \label{fig:sfig2}
\end{subfigure}
\caption{DANE combined with SGD and SVRG}
  \label{fig:sgd-svrg-all}
\end{figure}

In figure \ref{fig:sgd-svrg-all} we illustrate our result for using DANE combined with both SGD and SVRG. You can also see them compared with our baseline algorithms. We bring experiment with two different settings in terms of the number of samples and the machines, we experiment with $N = 6000, m=4$ and  $N = 12000 , m =16 $ and in terms of the number of the local iterations for SGD and SVRG we use $T \in \{ 0.5n , 2n\}$. The value of $T$ tells us how many stochastic gradient iterations are taken inside each iteration of DANE on one machine. As the first baseline algorithm we use SGD on a single machine and the same number of SGD iterations that we use for each iteration of DANE between any two cross points. We take distributed SGD as the second baseline, where we again use the same number of SGD iterations mentioned above between any two cross points. The difference of this one with the single machine SGD is that at each cross point the computing machines communicate and use the average of their iterates to continue with their local SGD. In order to get an idea on how well a distributed SGD can be in the ideal case we also plot a curve showing the first benchmark assuming that the cost is linearly decreased with the number of the machines. That can be obtained by dividing the number of the gradients to $4$ and $16$ as the corresponding number of gradients calculated on one machine in the distributed setting and keeping the same sub-optimality values.

We observe that DANE combined with SVRG converges linearly and outperforms DANE combined with SGD, distributed SGD and the optimal case imagined by distributing single machine SGD with no cost. Considering DANE combined by SGD on the other hand, we see that with low number of SGD steps per each DANE iteration, the algorithm does not show any advantage over distributed SGD as demonstrated on the plots on the left. However, using more SGD steps inside DANE iterations, DANE combined with SGD performs better than distributed SGD in converging to the optimal solution. Although using SVRG outperforms SGD we can see that SGD gives better convergence speed at the few initial iterations. Therefore it is reasonable to advice using SGD when we only need moderate accuracy and strictly stick to SVRG for applications demanding higher accuracies.

% TODO: \subsection{Same job but using different number of machines}

%Hello
%\subsection{DANE+SGD: different amount of local computations per iteration}
%Hello
%\subsection{DANE+SGD vs. Dist-SGD}
%Hello
%%\subsection{DANE+SGD vs. Simple DANE+SVRG (Dist-SVRG)}
%%Hello
%%\subsection{DANE+SVRG vs. Dist-SVRG}
%%Hello
%
%here we experiment with different parameters of $\mu$ and $\eta$ to see whether they can make any positive affect.
%
\subsection{Experiments with limited data access}

In this section we study the possibility of using partial data for our approximations of DANE. Studying this aspect is  useful for applications where we have very large number of samples. It is also useful for streaming applications or for online learning. We can consider two main parts where we can think of reducing the need to use all samples. The first one is related to calculating the gradient using all samples to form the DANE subproblems for each iteration. The second one is sampling from data on the local machines and using them for the stochastic steps inside SGD or SVRG. We specified two different versions of our algorithm trying to access partial data and see how they behave. In the following two subsections We provide experiments with these two versions.

\subsubsection{DANE combined with inexact solver and partial data}

\begin{figure}[ht]
\begin{subfigure}{.5\textwidth}
  \centering
  \includegraphics[width=1.\linewidth]{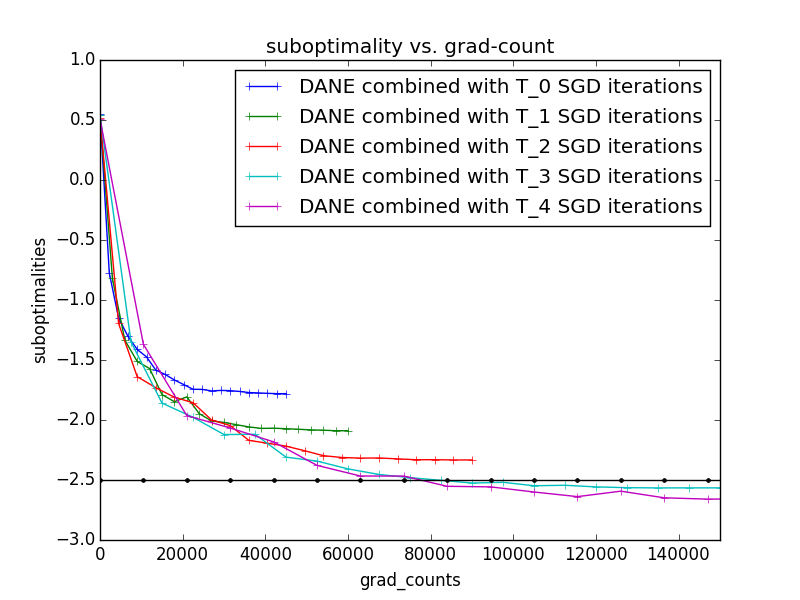}
  \caption{\scriptsize $(N , m) = (6000, 4)$, using 100\% of samples.\normalsize}   
  \label{fig:sfig1}
\end{subfigure}%
\begin{subfigure}{.5\textwidth}
  \centering
  \includegraphics[width=1.\linewidth]{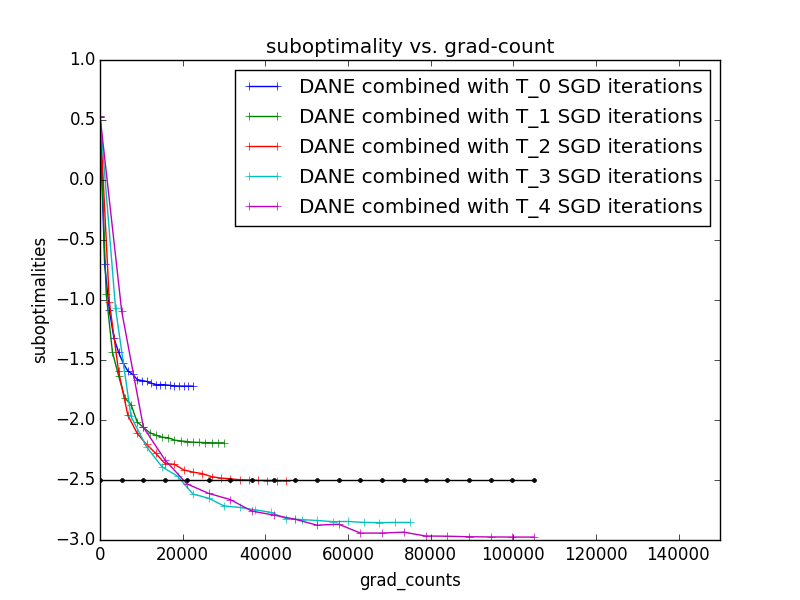}
  \caption{\scriptsize $(N , m) = (12000, 16)$, using 100\% of samples.\normalsize}   
  \label{fig:sfig2}
\end{subfigure} 
\begin{subfigure}{.5\textwidth}
  \centering
    \includegraphics[width=1.\linewidth]{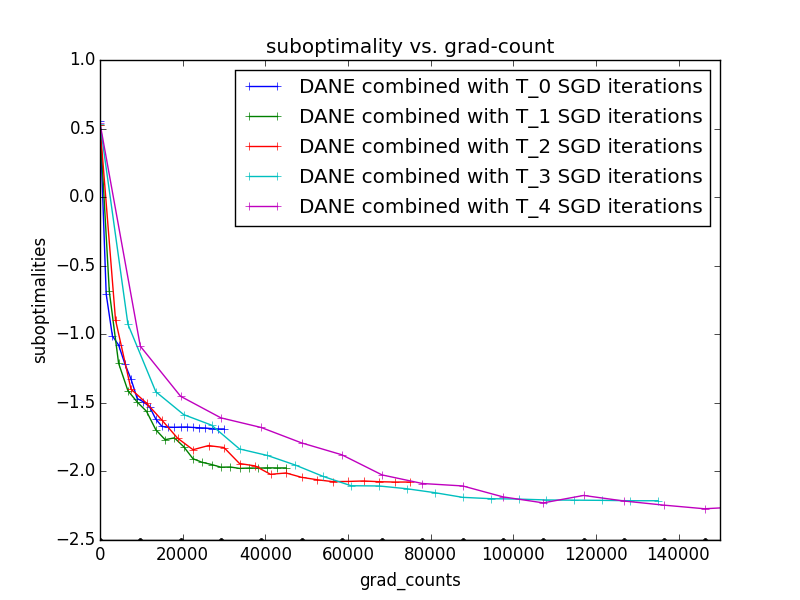}
  \caption{\scriptsize $(N , m) = (6000, 4)$, using 50\% of samples. \normalsize}   
  \label{fig:sfig2}
\end{subfigure}
\begin{subfigure}{.5\textwidth}
  \centering
   \includegraphics[width=1.\linewidth]{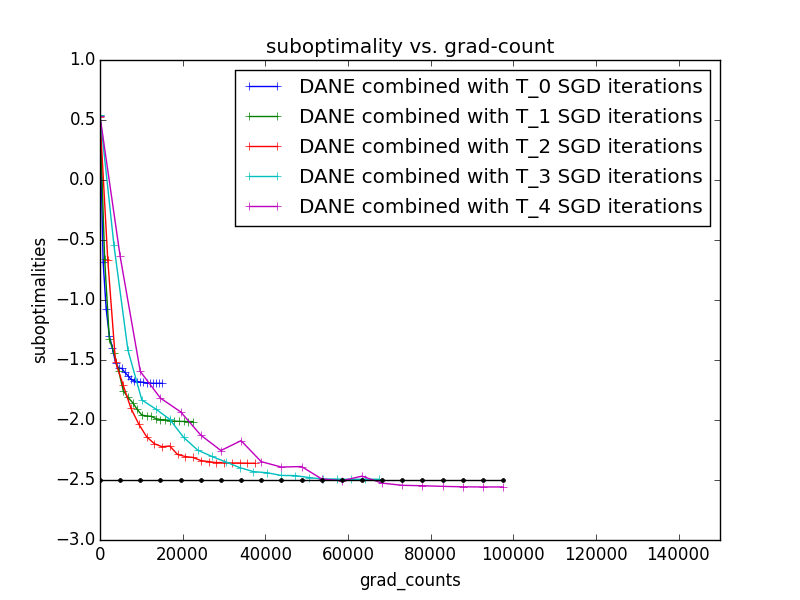}
  \caption{\scriptsize $(N , m) = (12000, 16)$, using 50\% of samples.\normalsize}   
  \label{fig:sfig2}
  
\end{subfigure}
\begin{subfigure}{.5\textwidth}
  \centering
  \includegraphics[width=1.\linewidth]{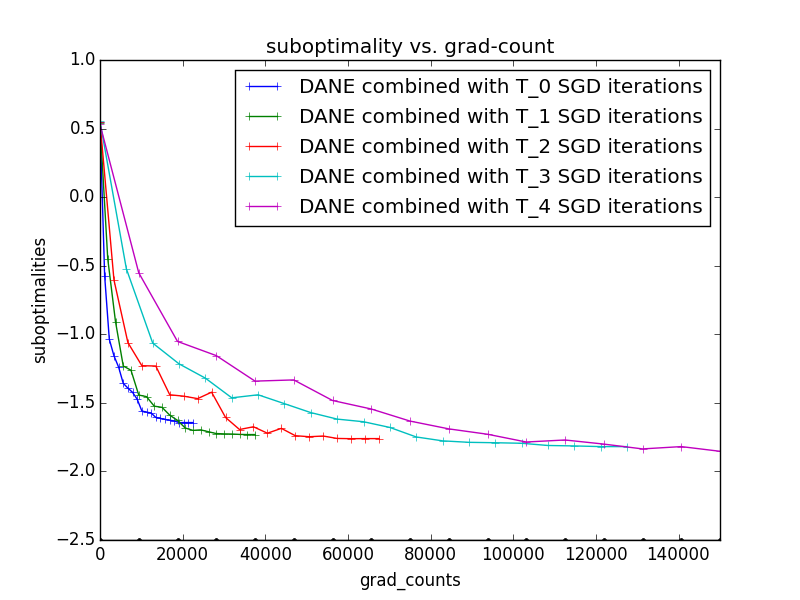}
  \caption{\scriptsize $(N , m) = (6000, 4)$, using 25\% of samples.\normalsize}   
  \label{fig:sfig2}
\end{subfigure}
\begin{subfigure}{.5\textwidth}
  \centering
  \includegraphics[width=1.\linewidth]{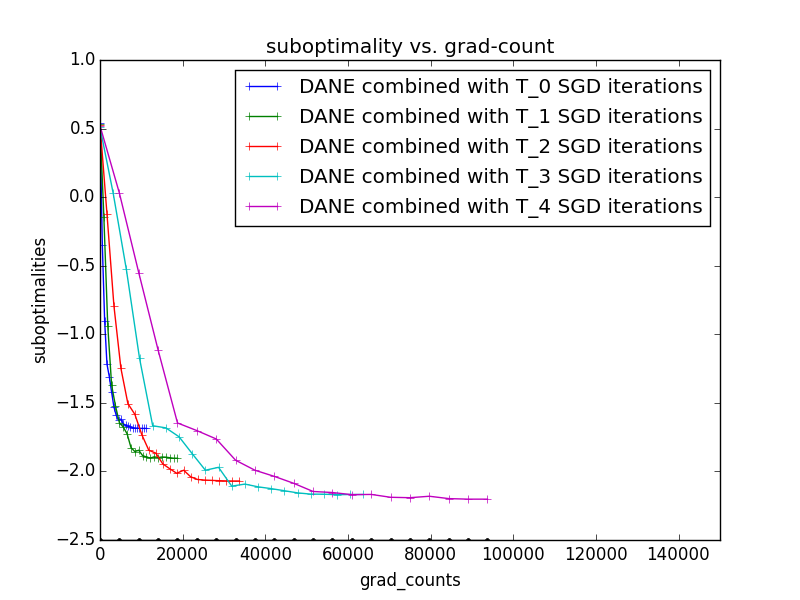}
  \caption{\scriptsize $(N , m) = (12000, 16)$, using 25\% of samples. \normalsize}   
  \label{fig:sfig2}
\end{subfigure}
\caption{DANE combined with SGD - access to fixed subset of samples.}
  \label{fig:dane-sgd-limited1}
\end{figure}

As our first limited access version, we assume limited access to only a subset of data on each machine at each iteration of DANE. For each DANE iteration we can only use a fixed subset of size $xn$ of the samples on each machine, with $0 < x \leq 1$. This subset is chosen randomly and is used for calculating the full gradients, as well as the stochastic steps taken inside SGD or SVRG. Figure \ref{fig:dane-sgd-limited1} shows the results of this approach for DANE combined with SGD. In the top row, all data is used whereas in the next rows only 50\% and 25\% of samples are used.

By going down on the rows of Figure \ref{fig:dane-sgd-limited1} and using partial data of smaller amount, we get lower accuracies from the same number of DANE iterations. However, the decrease in accuracy is not too big and this would be still a reasonable approach for many applications. Using a small number of SGD steps $T$ for each DANE iteration is not very useful in gaining moderately accurate solutions and we need as much as a few epochs for getting good solutions. On the other hand, from the computational point of view it might not seem reasonable to use a portion of data to decrease the computational cost while we are anyways computing on a few epochs of total data for each DANE iteration. However, this partial data access policy can be still helpful since reading from a smaller portion of data for each iteration improves the spacial locality of the instructions and results in lower cost for reading data from the memory.

\begin{figure}[htbp]
\begin{subfigure}{.5\textwidth}
  \centering
  \includegraphics[width=1.\linewidth]{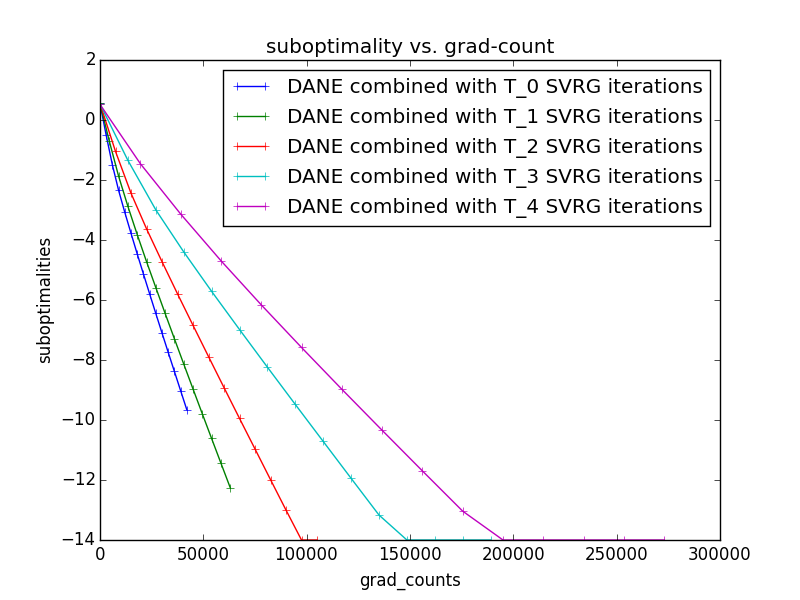}
  \caption{\scriptsize $(N , m) = (6000, 4)$, using 100\% of samples.\normalsize}   
  \label{fig:sfig1}
\end{subfigure}%
\begin{subfigure}{.5\textwidth}
  \centering
  \includegraphics[width=1.\linewidth]{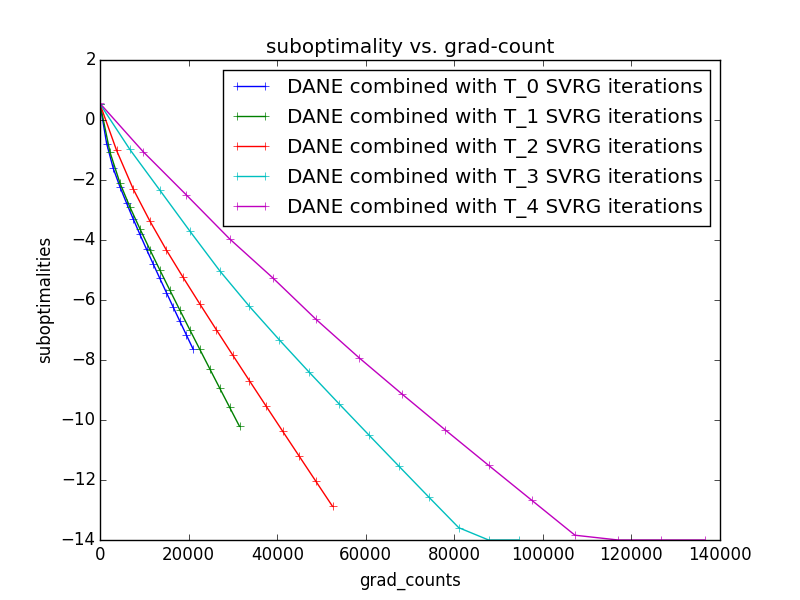}
  \caption{\scriptsize $(N , m) = (12000, 16)$, using 100\% of samples.\normalsize}   
  \label{fig:sfig2}
\end{subfigure} 
\begin{subfigure}{.5\textwidth}
  \centering
    \includegraphics[width=1.\linewidth]{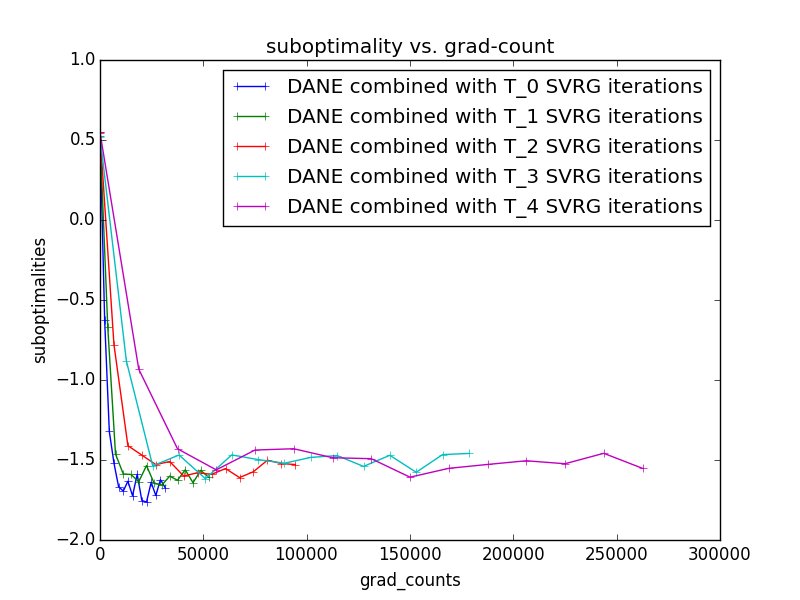}
  \caption{\scriptsize $(N , m) = (6000, 4)$, using 50\% of samples. \normalsize}   
  \label{fig:sfig2}
\end{subfigure}
\begin{subfigure}{.5\textwidth}
  \centering
   \includegraphics[width=1.\linewidth]{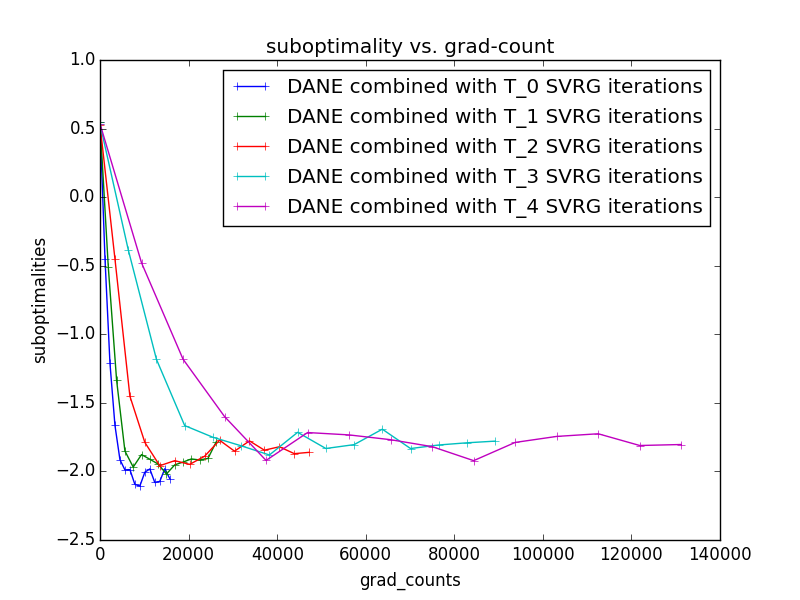}
  \caption{\scriptsize $(N , m) = (12000, 16)$, using 50\% of samples.\normalsize}   
  \label{fig:sfig2}
  
\end{subfigure}
\begin{subfigure}{.5\textwidth}
  \centering
  \includegraphics[width=1.\linewidth]{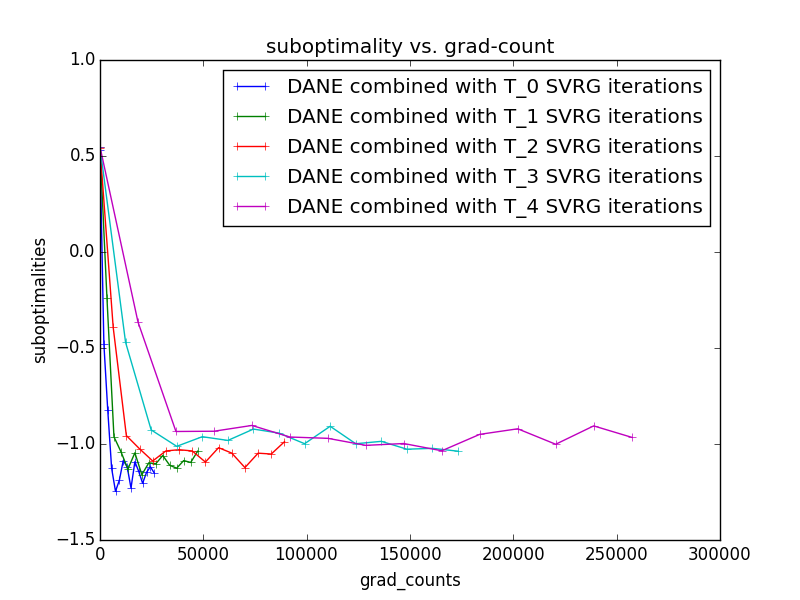}
  \caption{\scriptsize $(N , m) = (6000, 4)$, using 25\% of samples.\normalsize}   
  \label{fig:sfig2}
\end{subfigure}
\begin{subfigure}{.5\textwidth}
  \centering
  \includegraphics[width=1.\linewidth]{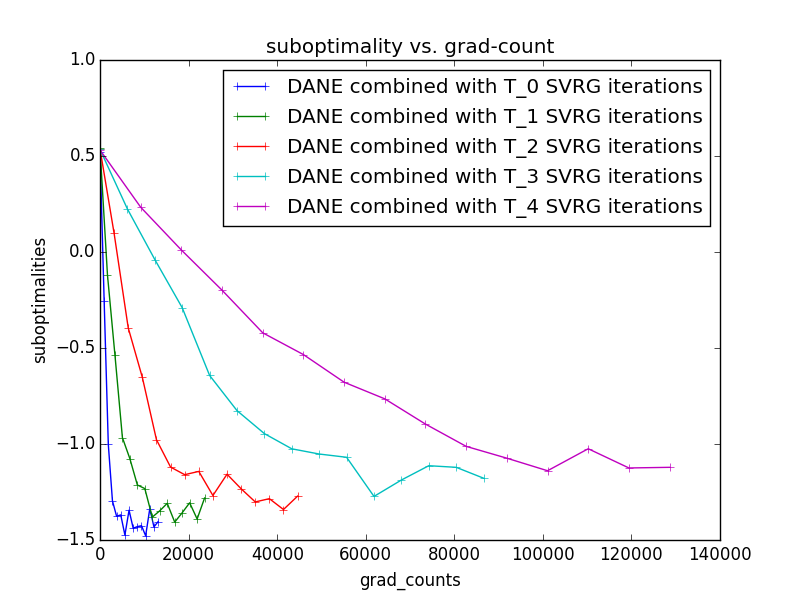}
  \caption{\scriptsize $(N , m) = (12000, 16)$, using 25\% of samples. \normalsize}   
  \label{fig:sfig2}
\end{subfigure}
\caption{DANE combined with SVRG - access to fixed subset of samples.}
  \label{fig:dane-svrg-limited1}
\end{figure}

We performed the same experiments with DANE combined with SVRG algorithm. Figure \ref{fig:dane-svrg-limited1} contains the results for this experiment. Again the top row shows the results when using the algorithm allowing it to access all samples. Looking at the second and third rows we can see that unlike the previous results for SGD, when we use only partial amount of data for SVRG, results become far worse than the results obtained when all samples were used. The results are even worse than what we get using DANE combined with SGD and similar set-up of the number of steps for each DANE iteration and the proportion of data used. Overall, using partial amount of data in this simple form when we are combining DANE with SVRG does not motivate using this algorithm for the scenarios where we have a lot of data and are not able to use all of them for solving the problem.

\subsubsection{DANE combined with inexact solver and approximate gradient}

As the second limited access version of our algorithms, we only use limited amount of data for calculation the gradients for each DANE iterations. The gradients on local machines and the full gradient used for each DANE iteration are then computed only a proportion of data which is chosen randomly for every DANE iteration. However, in contrast to the first version the stochastic steps inside SGD and SVRG can draw samples from all data-points on the local machines and are not limited to this set.

This version reduces the computation costs for calculating the gradient and is therefore useful when we do not plan to take a large number of stochastic steps in the local solvers and would like to balance the cost of calculating the gradient with that choice. It is also useful when we do not have access to all data at once when calculating the gradient but have access to infinite data to sample from fro running SGD and SVRG which is the case in streaming applications or online learning setting.

\begin{figure}[htbp]
\begin{subfigure}{.5\textwidth}
  \centering
  \includegraphics[width=1.\linewidth]{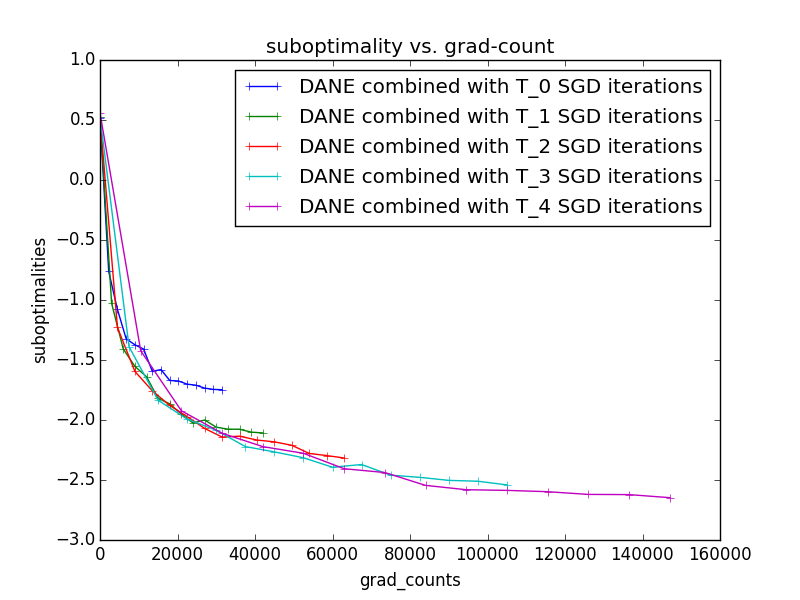}
  \caption{\scriptsize $(N , m) = (6000, 4)$, using 100\% of samples.\normalsize}   
  \label{fig:sfig1}
\end{subfigure}%
\begin{subfigure}{.5\textwidth}
  \centering
  \includegraphics[width=1.\linewidth]{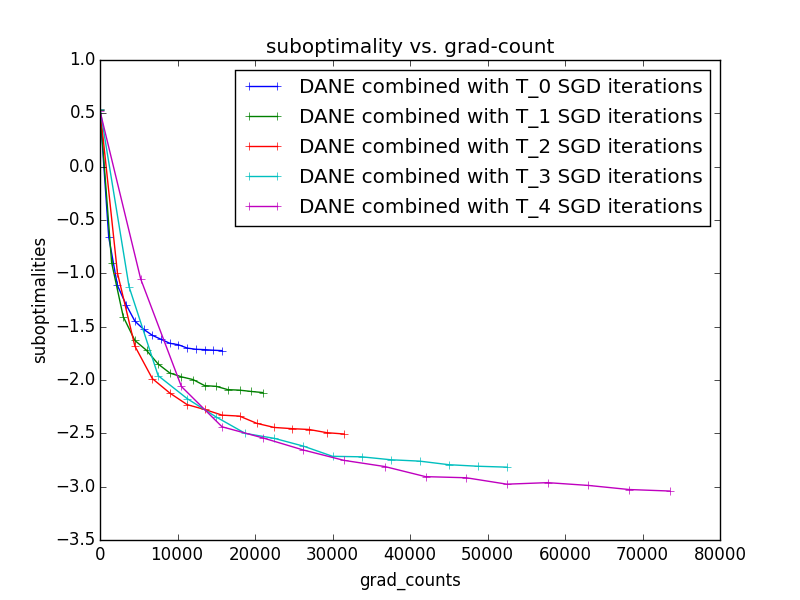}
  \caption{\scriptsize $(N , m) = (12000, 16)$, using 100\% of samples.\normalsize}   
  \label{fig:sfig2}
\end{subfigure} 
\begin{subfigure}{.5\textwidth}
  \centering
    \includegraphics[width=1.\linewidth]{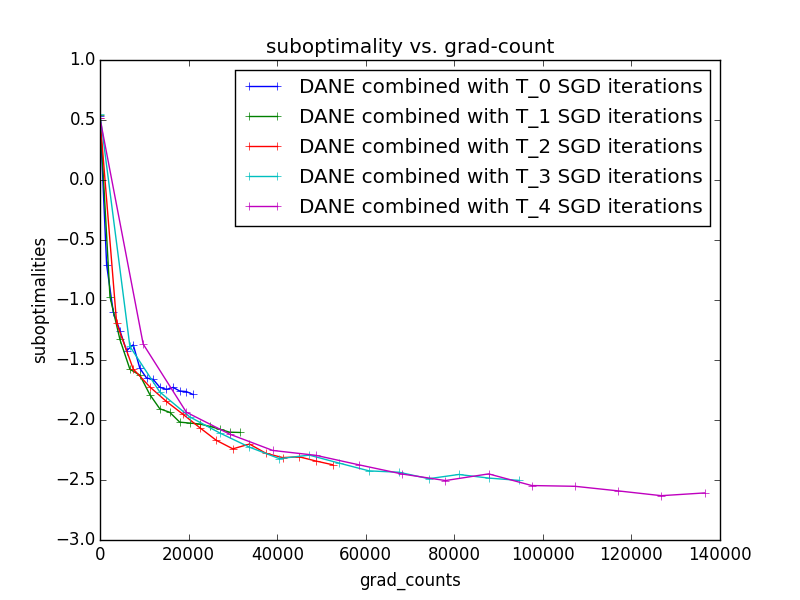}
  \caption{\scriptsize $(N , m) = (6000, 4)$, using 50\% of samples. \normalsize}   
  \label{fig:sfig2}
\end{subfigure}
\begin{subfigure}{.5\textwidth}
  \centering
   \includegraphics[width=1.\linewidth]{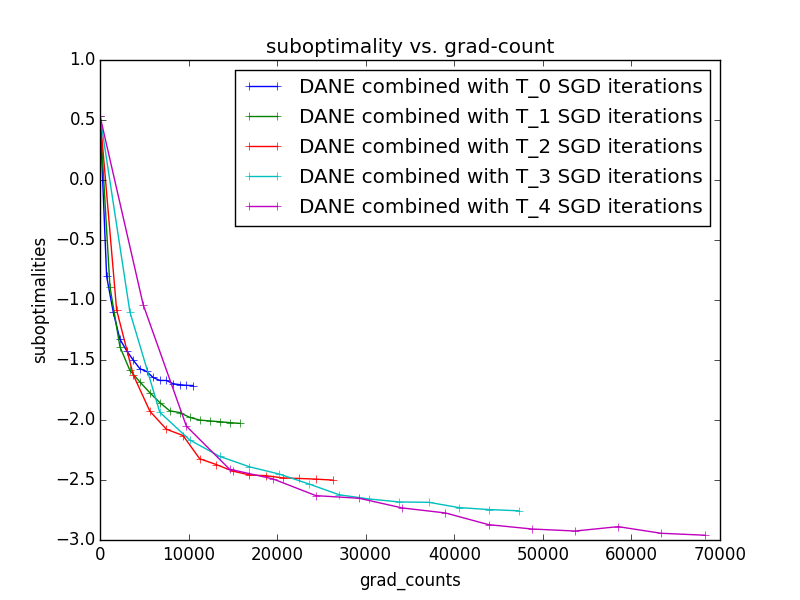}
  \caption{\scriptsize $(N , m) = (12000, 16)$, using 50\% of samples.\normalsize}   
  \label{fig:sfig2}
  
\end{subfigure}
\begin{subfigure}{.5\textwidth}
  \centering
  \includegraphics[width=1.\linewidth]{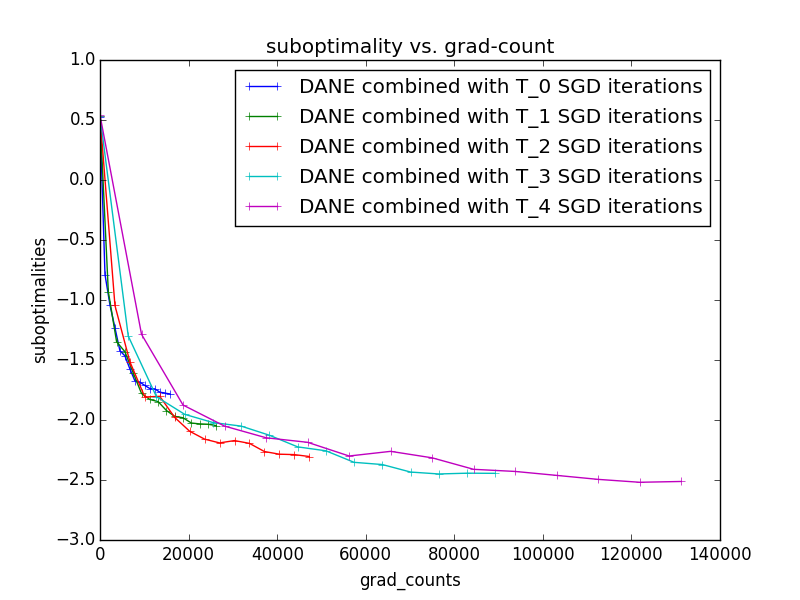}
  \caption{\scriptsize $(N , m) = (6000, 4)$, using 25\% of samples.\normalsize}   
  \label{fig:sfig2}
\end{subfigure}
\begin{subfigure}{.5\textwidth}
  \centering
  \includegraphics[width=1.\linewidth]{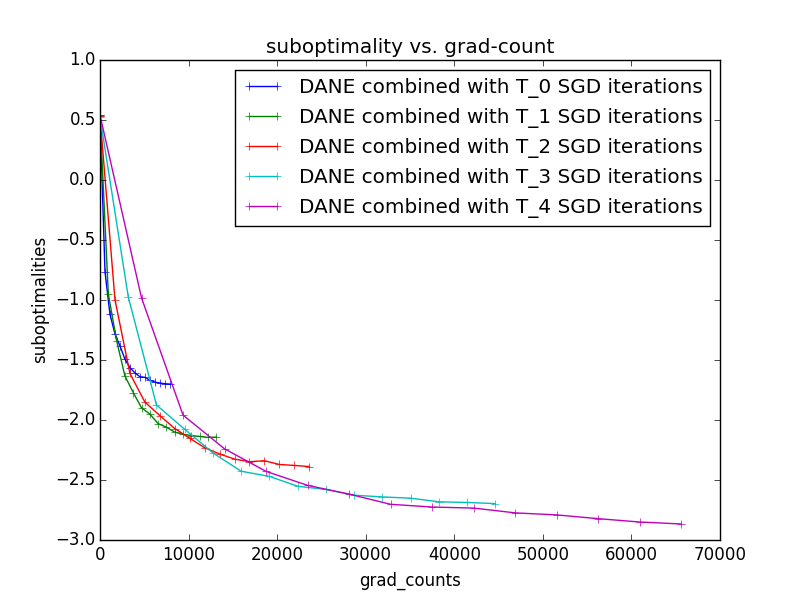}
  \caption{\scriptsize $(N , m) = (12000, 16)$, using 25\% of samples. \normalsize}   
  \label{fig:sfig2}
\end{subfigure}
\caption{DANE combined with SGD - global gradient calculated using a subset of samples.}
  \label{fig:dane-sgd-limited2}
\end{figure}

\begin{figure}[htbp]
\begin{subfigure}{.5\textwidth}
  \centering
  \includegraphics[width=1.\linewidth]{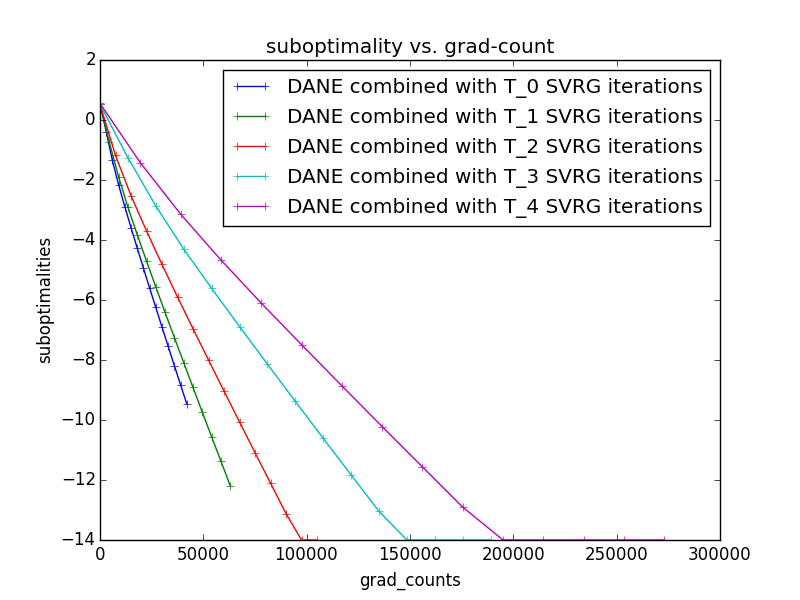}
  \caption{\scriptsize $(N , m) = (6000, 4)$, using 100\% of samples.\normalsize}   
  \label{fig:sfig1}
\end{subfigure}%
\begin{subfigure}{.5\textwidth}
  \centering
  \includegraphics[width=1.\linewidth]{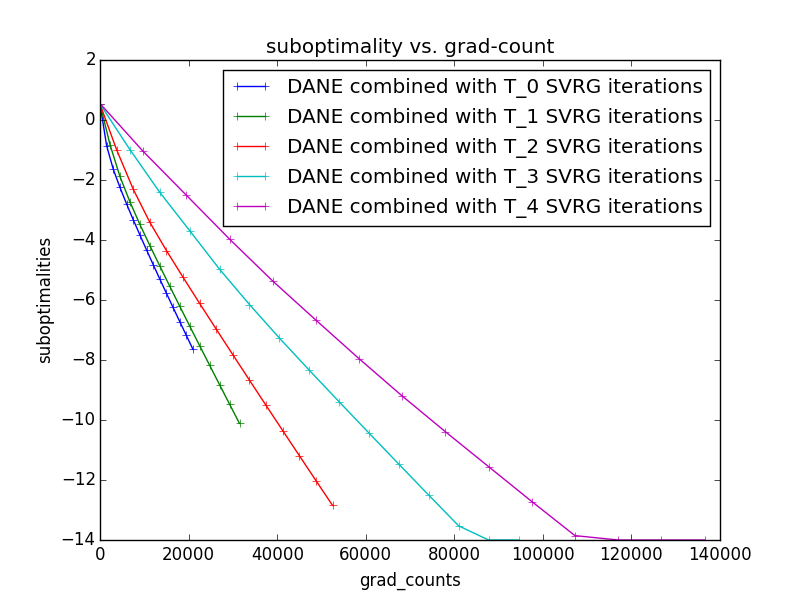}
  \caption{\scriptsize $(N , m) = (12000, 16)$, using 100\% of samples.\normalsize}   
  \label{fig:sfig2}
\end{subfigure} 
\begin{subfigure}{.5\textwidth}
  \centering
    \includegraphics[width=1.\linewidth]{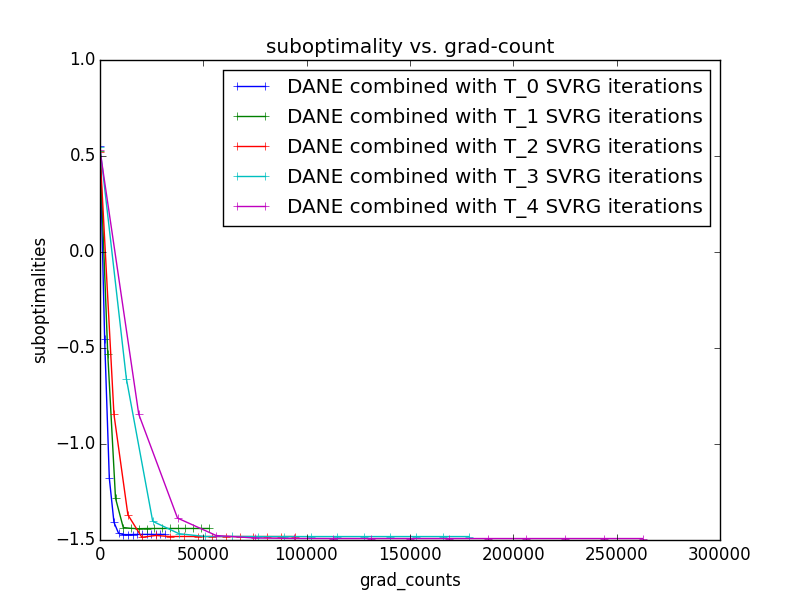}
  \caption{\scriptsize $(N , m) = (6000, 4)$, using 50\% of samples. \normalsize}   
  \label{fig:sfig2}
\end{subfigure}
\begin{subfigure}{.5\textwidth}
  \centering
   \includegraphics[width=1.\linewidth]{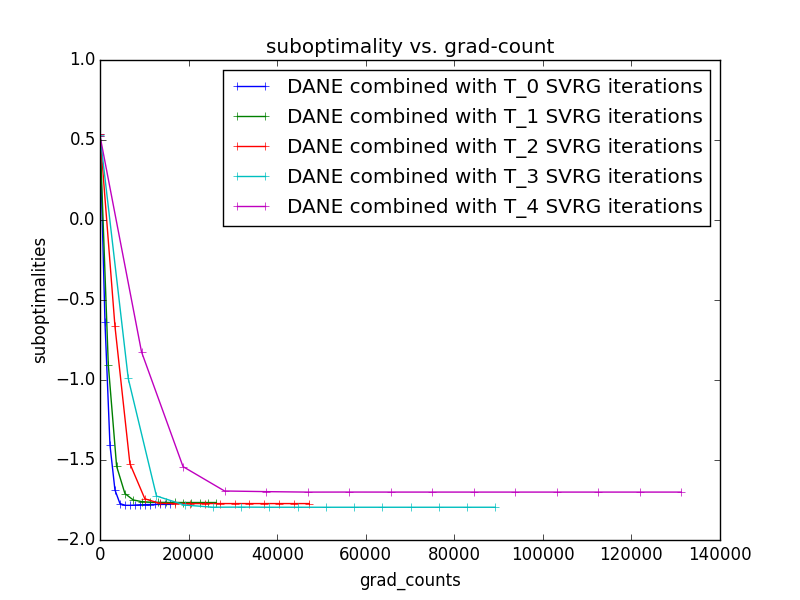}
  \caption{\scriptsize $(N , m) = (12000, 16)$, using 50\% of samples.\normalsize}   
  \label{fig:sfig2}
  
\end{subfigure}
\begin{subfigure}{.5\textwidth}
  \centering
  \includegraphics[width=1.\linewidth]{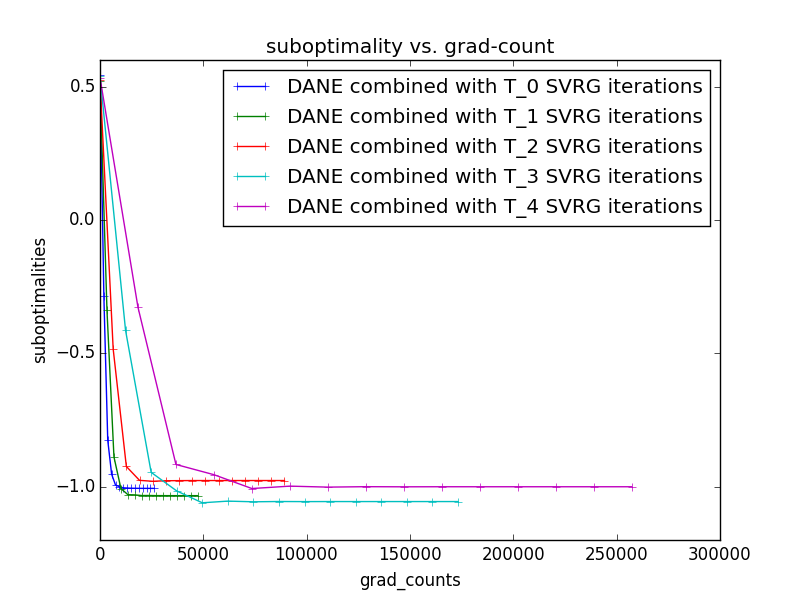}
  \caption{\scriptsize $(N , m) = (6000, 4)$, using 25\% of samples.\normalsize}   
  \label{fig:sfig2}
\end{subfigure}
\begin{subfigure}{.5\textwidth}
  \centering
  \includegraphics[width=1.\linewidth]{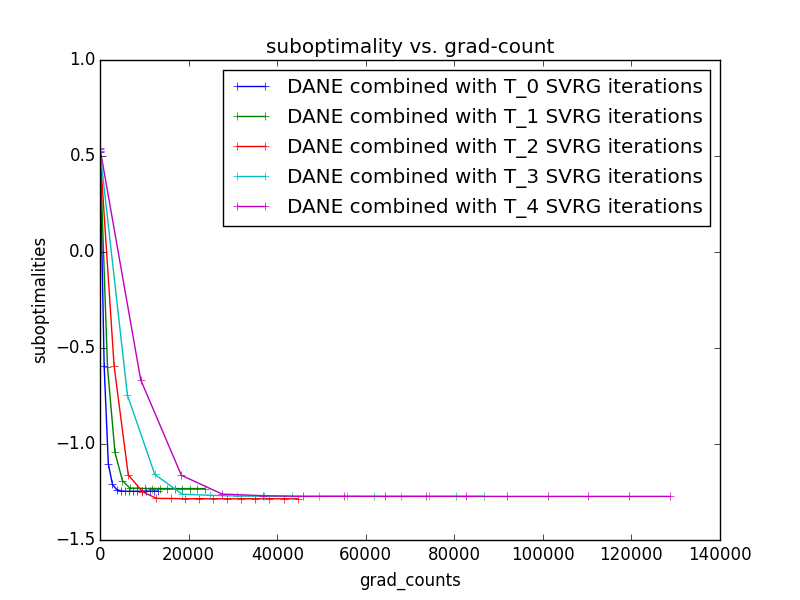}
  \caption{\scriptsize $(N , m) = (12000, 16)$, using 25\% of samples. \normalsize}   
  \label{fig:sfig2}
\end{subfigure}
\caption{DANE combined with SVRG - global gradient calculated using a subset of samples.}
  \label{fig:dane-svrg-limited2}
\end{figure}

We demonstrate the results for this version with SGD and SVRG in Figures \ref{fig:dane-sgd-limited2} and \ref{fig:dane-svrg-limited2}. The results for SGD in Figure \ref{fig:dane-sgd-limited2} are very interesting since they do not degrade much from the original version which uses full gradients calculated from all samples. Using this version slightly decreases the computation cost as the cost of the gradients is lower. However, unless if we use low number of SGD iterations this decrease in the cost would be negligible since the bigger cost comes from the SGD updates. However, it is very promising to use this algorithm for streaming scenarios when we can calculate the gradient using some recent samples and let the SGD steps use the data as they arrive for the updates.

The results for SVRG in Figure \ref{fig:dane-svrg-limited2} are again not as good as SGD and they become a lot worse than the original algorithms with exact gradient calculation. Here the results are a bit smoother than the results of our first version but do not get us to any higher accuracies. As well as the first version here again there is nothing to justify using SVRG as opposed to SGD.

%%Hello
%%\subsection{Results when having arbitrary big amount of data - plots for fixed amount of computation but different data sizes}
%%Hello

%\section{Distributed SGD with partial data}
%Hello
%
%- Are the results consistent with when using the whole data?

\section{Conclusion and Future Work}
\label{conc}

In this thesis we tried to provide practical solutions for solving distributed optimization for machine learning problems which are efficient in the amount of communication and enjoy reasonable run-time. We implemented an approximation for the DANE algorithm by leveraging stochastic local solvers for solving DANE sub-problems on the local machines.  As result we get algorithms which are not computationally extensive as the original DANE algorithm and other approaches with similar computation cost and require few rounds of communication. We provided two algorithms one with combining SGD as the local solvers in DANE and the other one with SVRG. Using experimental studies we demonstrated the capabilities of each of these two algorithms and provided insight for practical uses.

Our experiments with DANE combined with SGD demonstrated the algorithm is useful when we seek solutions with moderate accuracy and it is very fast for reaching to such solutions. However, for more accurate solutions SGD is not a proper option as the local solver which is mostly due to the decaying step-size.  On the other hand from our experiments with DANE combined with SVRG we learnt that SVRG is in fact very effective as the local solver inside DANE. Given enough DANE iterations and SVRG steps this algorithms provide solutions with very high accuracy. In addition to the accurate solutions of using SVRG in this framework we do not pay the extra cost of SVRG related to calculating the full gradients since the full gradient is already calculated for each iteration of DANE and therefore using SVRG is not much more expensive than SGD when we use all samples.

We studied the possibility of sing partial data in these two algorithms for the applications where accessing all samples is not possible. Although choosing SVRG as the local solver provides better results than SGD, SGD is more robust when we use partial data. SVRG on the other hand  suffers too much under this condition and performs even worse than SGD accessing the same part of data.

For the future work we would like to continue our study for the scenarios with limited data access and streaming applications. Our simple solutions to adapt the SVRG solver has not been successful and SGD is not enough for obtaining accurate result. It is still a question whether we could try to combine the ideas in SVRG with another way of calculating the gradients in order to be compatible to streaming scenarios or let the algorithm access to limited amount of data and keep its good convergence behavior. We will try to investigate the possibility of combining other SVRG alternative like SAGA with our algorithms and study their performance.



% \bibliography{iclr2019_conference}
% \bibliographystyle{iclr2019_conference}

\bibliography{optimization}
\bibliographystyle{iclr2019_conference}

\end{document}